\documentclass[manuscript,nonacm]{acmart} %,review,anonymous
\AtBeginDocument{%
  }

\usepackage[shortcuts,acronym]{glossaries}
\usepackage{tikz}
\usetikzlibrary{shapes.geometric, arrows, positioning}
\usepackage[table,dvipsnames]{xcolor}
\usepackage{pgfplots}
\usepackage{cleveref}
\graphicspath{{figs/}}
\DeclareGraphicsExtensions{.pdf,.jpeg,.png}
\usepackage{multirow}
\usepackage{subcaption}
\usepackage{url}

\begin{document}

\title[AuraMask]{AuraMask: An Extensible Pipeline for Creating User Acceptable Anti-Facial Recognition Image Filters}

%%%%%%%%%%%%%%%% Authors' Info %%%%%%%%%%%%%%%%%

\author{Jacob Lagogiannis}
\orcid{0000-0001-5777-6078}
\affiliation{%
    \institution{Franklin and Marshall College}
    \city{Lancaster}
    \state{PA}
    \country{United States of America}
}
\additionalaffiliation{%
    \institution{Georgia Institute of Technology}
    \city{Atlanta}
    \state{GA}
    \country{United States of America}
}
\additionalaffiliation{%
    \institution{Carnegie Mellon University}
    \city{Pittsburgh}
    \state{PA}
    \country{United States of America}
}
\email{jacob.logas@fandm.edu}

\author{William Agnew}
\affiliation{%
    \institution{Carnegie Mellon University}
    \city{Pittsburgh}
    \state{PA}
    \country{United States of America}
}

\author{Rosa I. Arriaga}
\orcid{0000-0002-8642-7245}
\affiliation{%
    \institution{Georgia Institute of Technology}
    \city{Atlanta}
    \state{GA}
    \country{United States of America}
}

\author{Sauvik Das}
\orcid{0000-0002-9073-8054}
\affiliation{%
    \institution{Carnegie Mellon University}
    \city{Pittsburgh}
    \state{PA}
    \country{United States of America}
}


\renewcommand{\shortauthors}{Lagogiannis et al.}

\begin{abstract}
Anti-facial recognition (AFR) image filters alter images in ways that are subtle to people but blinding to computer vision.
Yet, despite widespread interest in these technologies to subvert surveillance, users rarely use them in practice --- because the ``subtle'' alterations are visible enough to conflict with users' self-presentation goals.
To address this challenge, we propose AuraMask: a novel approach to creating AFR filters that are both adversarially effective and aesthetically acceptable. 
Using AuraMask, we produce 40 ``aesthetic'' filters that emulate popular ``one-click'' Instagram image filters.
We show that AuraMask filters meet or exceed the adversarial effectiveness of prior methods against open-source facial recognition models.
Moreover, in a controlled online user study ($N=630$) we confirm these filters achieve significantly higher user acceptance than prior methods.
Lastly, we provide our AFR pipeline to the community for accelerated research in adversarially effective and aesthetically acceptable protections.
\end{abstract}




\maketitle

\newacronym{ai}{AI}{Artificial Intelligence}
\newacronym{afr}{AFR}{Anti-Facial Recognition}
\newacronym{osn}{OSN}{Online Social Network}
\newacronym{cv}{CV}{Computer Vision}
\newacronym{ml}{ML}{Machine Learning}
\newacronym{aml}{AML}{Adversarial Machine Learning}
\newacronym{atn}{ATN}{Adversarial Transformation Network}
\newacronym{sai}{SAI}{Subversive AI}
\newacronym{compp}{CompP}{Computational Photography}
\newacronym{saia8}{SAIA-8}{Subversive AI Assessment}
\newacronym{depo}{DePO}{Decentralized Privacy Overlay}
\newacronym{pgd}{PGD}{Projected Gradient Descent}
\newacronym{lpips}{LPIPS}{Learned Perceptual Image Patch Similarity}
\newacronym{mtl}{MTL}{Multi-Task Learning}
\newacronym{gan}{GAN}{Generative Adversarial Network}
\newacronym{anova}{ANOVA}{Analysis of Variance}
\newacronym{poc}{PoC}{Proof of Concept}
\newacronym{sandp}{S\&P}{Security and Privacy}
\newacronym{usandp}{US\&P}{Usable Security and Privacy}
\newacronym{pet}{PET}{Privacy Enhancing Technology}
\newacronym{bipoc}{BIPOC}{Black, Indigenous, (and) People of Color}
\newacronym{lgbt}{LGBTQ+}{Lesbian, Gay, Bisexual, Transgender, and Queer}
\newacronym{ipfs}{IPFS}{Inter-Planetary File System}
\newacronym{hai}{HAI}{Human-centered AI}
\newacronym{hci}{HCI}{Human Computer Interaction}
\newacronym{fdf}{FDF}{Flickr Diverse Faces}
\newacronym{lfw}{LFW}{Labeled Faces in the Wild}
\newacronym{cld}{CLD}{Compact Letter Display}
\newacronym{ssim}{SSIM}{Structural Similarity Index Structure}
\newacronym{mae}{MAE}{Mean Absolute Error}
\newacronym{mse}{MSE}{Mean Squared Error}
\newacronym{sok}{SoK}{Systemization of Knowledge}
\newacronym{fe}{$\mathcal{L}_{FE}$}{Face Embedding}
\newacronym{fea}{$\mathcal{L}_{FEA}$}{Face Embedding Absolute}
\newacronym{fet}{$\mathcal{L}_{FET}$}{Face Embedding Threshold}
\newacronym{feat}{$\mathcal{L}_{FEAT}$}{Face Embedding Absolute Threshold}
\newacronym{fgsm}{FGSM}{Fast Gradient Sign Method}
\newacronym{mtcnn}{MTCNN}{Multi-Task Cascade Convolutional Networks}
\newacronym{cscw}{CSCW}{Computer-Supported Collaborative Work}
\newacronym{cmc}{CMC}{Computer-Mediated Communication}
\newacronym{cfa}{CFA}{Confirmatory Factor Analysis}
\newacronym{efa}{EFA}{Exploratory Factor Analysis}
\newacronym{vpn}{VPN}{Virtual Private Network}
\newacronym{dce}{DCE}{Deep Curve Estimation}
\newacronym{uve}{UVE}{Uncanny Valley Effect}
\newacronym{iuipc}{IUIPC}{Internet Users' Information Privacy Concern}
\newacronym{cid}{CID}{Content Identifier}
\newacronym{dct}{DCT}{Discrete Cosine Transform}
\newacronym{yass}{YASS}{Yet Another Steganographic Scheme \cite{kaushalsolanki_yass_2007}}
\newacronym{wysiwyg}{WYSIWYG}{What You See is What You Get}
\newacronym{pca}{PCA}{Principal Component Analysis}
\newacronym{llm}{LLM}{Large Language Models}
\newacronym{psnr}{PSNR}{Peak Signal-Noise Ratio}
\newacronym{fid}{FID}{Frechet Inception Distance}
\newacronym{mmd}{MMD}{Maximum Mean Discrepancy}
\newacronym{tipim}{TIP-IM}{Targeted Identity-Protection Iterative Method}
\newacronym{mim}{MIM}{Momentum Iterative Gradient-based Methods}
\newacronym{mtadv}{MTADV}{Multi-Task Adversarial Attack}
\newacronym{nima}{NIMA}{Neural Image Assessment}
\newacronym{singletarget}{ST}{Single-Target Defense}
\newacronym{ensembletarget}{ET}{Ensemble-Target Defense}

\begin{figure*}[h]
  \caption*{\acrfull{singletarget}}
  \begin{subfigure}{0.12\linewidth}
    \includegraphics[width=\textwidth]{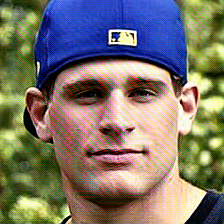}
    \caption{Dogpatch}
    \label{fig:dogpatch}
  \end{subfigure}
  \begin{subfigure}{0.12\linewidth}
    \includegraphics[width=\textwidth]{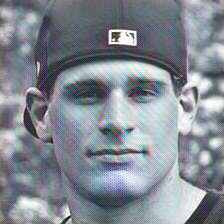}
    \caption{Moon}
    \label{fig:moon}
  \end{subfigure}
  \begin{subfigure}{0.12\linewidth}
    \includegraphics[width=\textwidth]{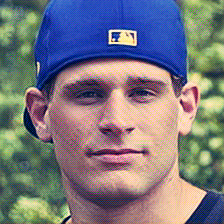}
    \caption{Nashville}
    \label{fig:nashville}
  \end{subfigure}
  \begin{subfigure}{0.12\linewidth}
    \includegraphics[width=\textwidth]{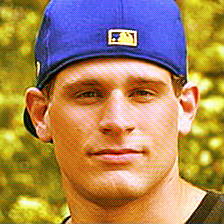}
    \caption{Kelvin}
    \label{fig:kelvin}
  \end{subfigure}
  \begin{subfigure}{0.12\linewidth}
    \includegraphics[width=\textwidth]{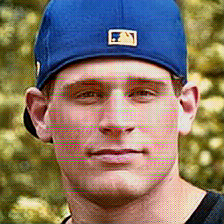}
    \caption{Sierra}
    \label{fig:sierra}
  \end{subfigure}
  \begin{subfigure}{0.12\linewidth}
    \includegraphics[width=\textwidth]{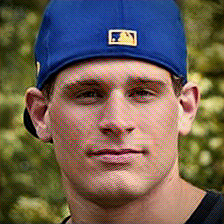}
    \caption{Sutro}
    \label{fig:sutro}
  \end{subfigure}
  \begin{subfigure}{0.12\linewidth}
    \includegraphics[width=\textwidth]{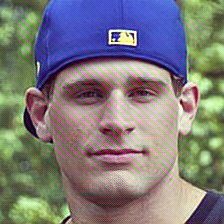}
    \caption{Valencia}
    \label{fig:valencia}
  \end{subfigure}

  \caption*{\acrfull{ensembletarget}}
  \begin{subfigure}{0.12\linewidth}
    \includegraphics[width=\textwidth]{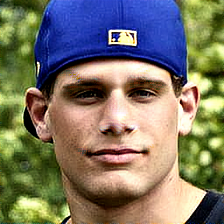}
    \caption{Dogpatch}
    \label{fig:dogpatch_ensemble}
  \end{subfigure}
  \begin{subfigure}{0.12\linewidth}
    \includegraphics[width=\textwidth]{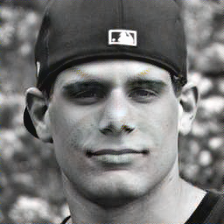}
    \caption{Moon}
    \label{fig:moon_ensemble}
  \end{subfigure}
  \begin{subfigure}{0.12\linewidth}
    \includegraphics[width=\textwidth]{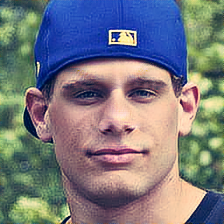}
    \caption{Nashville}
    \label{fig:nashville_ensemble}
  \end{subfigure}
  \begin{subfigure}{0.12\linewidth}
    \includegraphics[width=\textwidth]{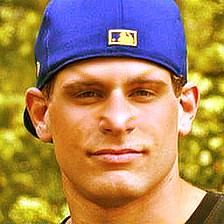}
    \caption{Kelvin}
    \label{fig:kelvin_ensemble}
  \end{subfigure}
  \begin{subfigure}{0.12\linewidth}
    \includegraphics[width=\textwidth]{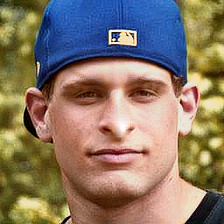}
    \caption{Sierra}
    \label{fig:sierra_ensemble}
  \end{subfigure}
  \begin{subfigure}{0.12\linewidth}
    \includegraphics[width=\textwidth]{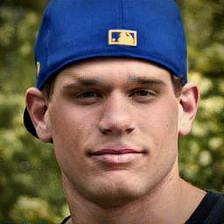}
    \caption{Sutro}
    \label{fig:sutro_ensemble}
  \end{subfigure}
  \begin{subfigure}{0.12\linewidth}
    \includegraphics[width=\textwidth]{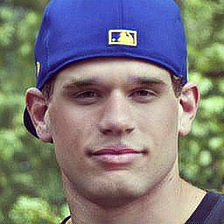}
    \caption{Valencia}
    \label{fig:valencia_ensemble}
  \end{subfigure}
  \caption{Aesthetic \acrshort{afr} defenses with Single (top) and Ensemble (bottom) Targets produced with the AuraMask pipeline.}
  \label{fig:obfuscation-examples}
\end{figure*}

\section{Introduction}
% State of the world: 

% \begin{figure*}[tbh]
%   \centering
%   \includegraphics[width=\textwidth]{figs/Obfuscations/teaser-figure.png}
%   \caption{A set of 10 example Instagram-like anti-facial recognition filters produced by AuraMask.}
%   \label{fig:obfuscation-examples}
% \end{figure*}

Facial recognition has never been easier.
It is now possible to reliably and automatically identify people from photos or videos of their face even if partially occluded -- wearing a mask \cite{vu_masked_2022,zeng_survey_2021,ullah_realtime_2022} -- or captured in suboptimal environments -- low lighting \cite{fan_lowfacenet_2024}.
% Folks often use facial recognition to organize their personal photos, secure their devices, or keep an eye on their home \cite{mcclellan_facial_2019}.
Facial recognition is increasingly pervasive.
% Employers are using it to monitor and track employee movements, productivity, and even predicted emotional state \cite{boyd_automated_2023,stark_dont_2020}.
Governments use it to create a panopticon for effortless monitoring of the movements, associations, and behaviors of their population \cite{parvini_surveillance_2024,diaz_new_2019,mac_clearviews_2020,laidler_surveillance_2008,manokha_surveillance_2018,gray_urban_2003,selinger_inconsentability_2020}.
Facial recognition has never been more pernicious.
Employers employ facial recognition for ``algorithmic management'' to micromanage employee work and undermine unionization efforts \cite{hanley_eyes_2020,wiggin_weaponizing_2025}.
Police departments around the world use flawed facial recognition systems which has only reinforced racialized policing \cite{diaz_new_2019,scheuerman_how_2020,browne_dark_2015} and are regularly misused to identify and arrest political adversaries \cite{shepherd_artist_2020,biddle_police_2020,brandom_facebook_2016,mansoor_propalestinian_2025}.
% Now, regardless of laws against it, Immigration and Customs Enforcement (ICE) were given unlimited access to these systems and nationalized these harms; using it to uncover dissent \cite{parvini_surveillance_2024,mansoor_will_2025,mansoor_propalestinian_2025}, identify immigrants \cite{newman_ice_2025}, and track their movements \cite{koebler_ice_2025} so they may be detained in concentration camps \cite{aleman_what_2025,charalambous_its_2025}.
% Outside the US, the ``Red Wolf'' facial recognition-based surveillance system deployed by the Israeli government aids in Palestinian apartheid \cite{amnestyinternational_israeli_2023,asseburg_amnesty_2022,amnestyinternational_israels_2022}.
% It is unlikely that the state will willingly dispose this surveillance apparatus, nor will the corporations responsible end their collaboration \cite{frenkel_microsoft_2018,roberts_facebook_2011,ulu-laniboyanton_protesters_2025,guliani_amazon_2018}; thus,
As the journalist Kashmir Hill argues, modern facial recognition may very well mark the ``the end of privacy as we know it.'' \cite{hill_this_2020}

Yet, facial recognition is fragile.
The underlying models and methods are reliant on high-quality data collected by corporations from their users \cite{jones_ai_2024}.
This reliance introduces opportunities to avoid, sabotage, or otherwise undermine facial recognition systems.
\acrfull{afr} photo defenses \cite{chandrasekaran_faceoff_2021,hussain_reface_2023,shan_fawkes_2020,cherepanova_lowkey_2021,wenger_sok_2023} emerged as a method for resistance which introduce ``subtle'' perturbations into user-shared images that complicate learning (poison) or introduce doubt (evasion) into the facial recognition pipeline.
One technique, Fawkes, was featured in the New York Times and was downloaded nearly a million times \cite{hill_this_2020, shan_fawkes_2020} shortly after public release.
Despite initial public interest, however, few people regularly use these \acrshort{afr} defenses on photos they share; one reason being that adversarial methods tend to introduce visible artifacts into images do not align with users' self-presentation goals \cite{logas_subversive_2024}.

% Indeed, while these techniques aim for ``subtlety'' by minimizing the pixel distance between the filtered and original images \cite{shan_fawkes_2020}, the reality is that for these techniques to be practically effective at poisoning or evasion, they often introduce unappealing human-visible artifacts into photos (see Figure XX).
% Moreover, recent work has shown that many people do not accept anti-facial recognition filters that are imperceptible because they do not believe imperceptible modifications will be effective \cite{logas_subversive_2024}.
%Moreover, without introducing perceptible changes to their images, people simply don't believe these techniques are effective \cite{}.

Users have strong aesthetic and self-presentation preferences when sharing photos online \cite{logas_subversive_2024,monteiro_imago_2025} which often override secondary privacy concerns \cite{logas_image_2022,dourish_security_2004, krsek_measuring_2025,monteiro_imago_2025,hasan_viewer_2018}.
\acrshort{afr} defenses with visible artifacts were found to conflict with a user's self-presentation goals \cite{logas_subversive_2024} and are thus unlikely to have sustained use.
However, users often use perceptible image filters before sharing photos online for aesthetic or creative purposes \cite{bakhshi_why_2015,javornik_what_2022} and indeed many images are algorithmically processed at the moment of capture on mobile phones \cite{penousalmachado_experiments_2008,yang_iaacs_2023}.
Thus, we seek to understand (\textbf{RQ1}) ``How can we design anti-facial recognition defenses that are better aligned with people's aesthetic and self-presentation preferences?''
To investigate, we present the extensible \textsc{AuraMask} toolkit that supports \acrfull{mtl} \cite{zhao_multitask_2023} training of \acrshort{afr} defenses with aesthetic \textit{and} adversarial objectives.

% The social networks Instagram and Snapchat offer one-click affordances to apply graphic alterations to images before sharing for more than a decade.
% In addition, literature in computational aesthetics that aims to model what makes a good, high-quality photo and then generate alterations to make the photos more appealing \cite{penousalmachado_experiments_2008,yang_iaacs_2023}.
% Here, our key insight is that through \acrfull{mtl} \cite{zhao_multitask_2023}, we can create \acrshort{afr} defenses that achieve user acceptance \textit{and} disrupt the facial recognition pipeline.
% In this paper, we present tangible examples of this novel approach along with an extensible toolkit -- \textsc{AuraMask} -- to create ``aesthetic'' \acrshort{afr} defenses.

% TODO: Leaning towrad a more "sub-question" approach here. Could be useful.
However, neither user acceptance nor adversarial effectiveness necessarily follow from the proposed design, thus we pose two additional research questions:
\textbf{RQ2} How do aesthetic \acrshort{afr} defenses compare to existing defenses in evading facial recognition?
\textbf{RQ3} How do aesthetic \acrshort{afr} defenses compare to existing defenses in terms of user acceptance?

We address RQ2 by evaluating our aesthetic \acrshort{afr} defenses on a battery of facial recognition benchmarks using state-of-the-art recognition models.
Our evaluations demonstrated that our \textsc{AuraMask} defenses retained adversarial effectiveness consistent with or better than prior approaches (i.e., LowKey \cite{cherepanova_lowkey_2021} and Fawkes \cite{shan_fawkes_2020}, respectively).

To answer RQ3, we conducted a two-iteration user study with $N=630$ participants who gauged the acceptability of several \acrshort{afr} defenses on the \acrfull{saia8} scale \cite{logas_subversive_2024}, a psychometric scale developed to measure user acceptance of image obfuscation.
Through our analysis, participants found \textsc{the AuraMask} defenses to be significantly more acceptable and preferable to prior approaches.
% , and found that, when presented with our aesthetic filters, they would choose to use an AFR method Y\% of the time, compared with just Z\% of the time when presented with just Fawkes of LowKey.

In summary, our key contributions are:
\begin{itemize}
    \item A pipeline to generate \acrshort{afr} defenses that account for the output aesthetics.
    \item A set of 80 proof-of-concept \textsc{AuraMask} defenses that emulate existing (non-protective) image filters.
    \item Validation and comparison of technical efficacy in both ''white-box`` and ''black-box`` scenarios.
    \item Empirical evidence of greater user acceptance of \textsc{AuraMask} aesthetic \acrshort{afr} defenses over prior methods.
\end{itemize}

\subsection{Are Anti-Facial Recognition Defenses Ethical?}
There is a debate within the \acrlong{sandp} community on whether user-facing \acrshort{afr} technologies is appropriate to address facial recognition harms.
% We feel the need to address the current discourse within the \acrfull{sandp} community wherein approaches like AuraMask are critiqued as overly simplistic, if not outright dangerous.
The critiques raised (e.g., \cite{radiya-dixit_data_2021}) have merit and deserve careful consideration; we discuss these critiques and our own perspective in more depth in the Discussion section (\ref{sec:discussion_protection}).

\section{Background}
% Since the advent of the \acrfull{osn}, social computing researchers have sought to understand user behavior online.
% Engagement happens through many avenues, from microblogging -- e.g., Twitter, Bluesky, Mastodon -- to publishing live video to a largely anonymous audience -- e.g., Twitch.
\acrfull{osn}s, for all their benefits, have had a number of key negative social impacts \cite{brock_blackhand_2012, byron_hey_2019, lup_instagram_2015}.
One of these negative impacts is an increased risk of privacy invasions.
% Invasions which may be mitigated through \emph{active non-participation} -- the ``politically willful engagement in a platform in order to slow it down or disrupt it, or exiting the platform entirely due to active decision-making'' \cite{casemajor_nonparticipation_2015}.
To mitigate these risks, the \acrfull{sandp} community has long developed user-focused, privacy-enhancing technologies that enable ``active non-participation'' -- i.e., \emph{obfuscation}, \emph{sabotage}, and \emph{exodus} \cite{casemajor_nonparticipation_2015}.
Most of these tools, however effective, have low overall adoption.
Here, we give an overview of prior work related to how and why users share personal images online. We also present research on efforts to protect individual privacy when sharing said images.

\subsection{Photo Sharing Behaviors Online}
A variety of complex factors contribute to a persons' motivation to share personal information online \cite{tifentale_selfiecity_2015, amon_influencing_2020}.
Oeldorf-Hirsch and Sundar identified four motivating factors for online image sharing \cite{oeldorf-hirsch_social_2016}: Seeking and Showcasing Experiences, Technological Affordances, Social Connection, and Reaching Out.
In addition, much of the literature suggests that one's self presentation greatly influences \emph{what} is shared online \cite{davies_display_2007, proudfoot_saving_2018, hong_you_2020, fox_selective_2016, salomon_that_2021} as are social pressures \cite{hallam_online_2017, church_user_2020, taddicken_privacy_2014, hu_what_2014, bell_you_2019, hasan_your_2021}.
While seemingly not a primary concern, privacy also impacts what is considered shareable online influencing some to engage in actions to mitigate privacy invasions \cite{ahern_overexposed_2007,zhao_understanding_2022}.

\subsubsection{Image Alterations in Sharing}
Even before the rise of image-based \acrshort{osn}s (e.g., Instagram), people touched up their photos in line with their self-presentation preferences.
While there was a movement toward more authentic, unaltered photos on social media \cite{salisbury_nofilter_2017}, people still regularly modify personal images before sharing \cite{bakhshi_why_2015}.
The popularity of this behavior can be attributed, in part, to the breadth of available alterations which introduced a new dimension for creativity \cite{petrelli_family_2010,ibanez-sanchez_augmented_2022, peng_time_2017} and enhanced control over public presentation \cite{brewster_that_2025, javornik_what_2022, hong_you_2020}.
Moreover, altered images achieve a higher audience engagement \cite{vendemia_effects_2021,bakhshi_why_2015}, a benefit in the attention economy \cite{menczer_attention_2020, zulli_capitalizing_2018}.
It is therefore unsurprising that users expend time and effort in applying modifications to their images before sharing \cite{bakhshi_why_2015}.

\subsection{Active Non-Participation through Obfuscation}
Obfuscation is the deliberate addition of ambiguous, confusing, or misleading information to interfere with surveillance and data collection with the goals of buying time, providing cover or deniability, evading observation, interfering with profiling, or expressing protest \cite{brunton_obfuscation_2015}.
Obfuscation tactics have a long history of use by marginalized populations to communicate under the gaze of powerful centralized institutions.
%--- so much so that Brunton and Nissenbaum call obfuscation the ``weapon of the weak.''
However, efforts to avoid observation also animates forensic efforts to detect and de-obfuscate secret messages.
This kind of ``arms race'' is familiar to those within the security community \cite{li_survey_2011, radiya-dixit_data_2021, vuyyuru_biologically_2020}.
Thus, it is important to note that the goal of obfuscation is to evade, obstruct, or buy time, and not necessarily to \emph{permanently} protect.
Image obfuscation can take many forms, but here we focus on image obfuscation which can be \textbf{interactive} or \textbf{authored}.

\subsubsection{Interactive Image Obfuscation} \label{sec:background-interactive-obfuscation}
This form of obfuscation requires interaction by both parties -- image sharer and observer -- and allows for greater affordances in secret data access control.
Oftentimes, this obfuscation approach requires splitting the image into parts that are encrypted \cite{he_puppies_2016} or stored elsewhere \cite{ra_p3_2013}.
Others use techniques like steganography to hide secret data in plain sight \cite{logas_image_2022,shumeetbaluja_hiding_2017,amsden_transmitting_2014}.
% Ra et al. obfuscate sensitive images by splitting the image into two parts, one public which is posted to social media and the other which is saved in a cloud provider -- e.g. Dropbox -- with the two pieces only combined if a viewer has to appropriate permissions \cite{ra_p3_2013}.
% He et al. similarly encrypted only portions of an image -- e.g., a face -- to be decrypted by the key-holder \cite{he_puppies_2016}.
These approaches allow for granular protection of shared images; however, their use is often simple to detect by a casual observer as the publicly shared images are highly perturbed and thus simple to disrupt by a central authority.

% \subsubsection{Steganographic Obfuscation}
% Steganography is an interactive obfuscation approach wherein data is hidden within a ``cover'' medium -- e.g., photo, document, or video.
% The effectiveness of steganography is determined by three heuristics: data retention, data detect-ability, and data capacity.
% With FaceCloak, Luo et al. obfuscated all text fields on Facebook, hot-swapping the fake values with real ones from a proprietary server \cite{luo_facecloak_2009}.
% Directly related to FaceCloak, NOYB protects Facebook user profile data by shuffling ``atoms'' of data among NOYB users and disentangling them by reference via a public dictionary file -- salted with non-user data to hinder mining by adversary \cite{guha_noyb_2008}.
% Hummingbird by Cristofaro et al. provided privacy on Twitter by running a parallel service to encrypt a users tweets and decrypt them for approved followers \cite{cristofaro_hummingbird_2012}.
% Lucas and Borisov introduced FlyByNight, a Facebook application for the encryption of direct messages \cite{lucas_flybynight_2008}; accomplished by registering as a Facebook app and running a proprietary server.
% Beato et al. presents a tool for hiding links to data on a publicly addressable storage service; however, this scheme still relies on third-party ``hash-map services'' -- e.g., TinyURL -- for content delivery \cite{beato_eyes_2013}.
% In Chapter \ref{ch:imagedepo}, we propose a novel system that steganographically obfuscates private images from institutions without necessitating a trusted third party.

\subsubsection{Authored Image Obfuscation}\label{sec:background-authored-obfuscation}
Authored image obfuscation only requires action on the part of the sharer who modifies the secret data in a way that is unintelligible to a specific threat but is understood by the intended audience.

\subsubsection{Obfuscation through Redaction}
The most straightforward way to avoid sensitive data exposure is through redaction.
Hassan et al. automated redaction with a tool that autonomously classified sensitive image portions and obscured them with a cartoon image \cite{hassan_cartooning_2017}.
% In a qualitative evaluation of their tool, Hasan et al. found that users make a trade-off between the aesthetics of an obfuscated image and the protection it provides \cite{hasanViewerExperienceObscuring2018}.
Other techniques target face recognition by blurring \cite{li_effectiveness_2017}, cartooning \cite{erdelyi_serious_2013, hassan_cartooning_2017}, or otherwise redacting a visible face.
As with interactive methods, a drawback of redaction is noticeability as the changes are immediately evident and could negatively impact image utility.
However, advances in \acrfull{aml} introduced an opportunity to mitigate this drawback by requiring only slight data alterations to sabotage deep learning models \cite{szegedy_intriguing_2014}.

\subsubsection{Adversarial Obfuscation}
Modern face obfuscation systems can sabotage facial recognition with adversarial image alterations \cite{shan_fawkes_2020,cherepanova_lowkey_2021,ilia_face_2015,hussain_reface_2023, wenger_sok_2023}.
Using \acrshort{aml} techniques \cite{szegedy_intriguing_2014}, these systems generate adversarial noise to fool the underlying processes of the facial recognition pipeline.
Within \acrshort{afr}, multiple stages of the facial recognition process can be targeted -- i.e., image capture, processing, feature extraction, reference database creation, or database querying for identification \cite{wenger_sok_2023}.
Although prior work has proposed a variety of effective \acrshort{afr} methods, the approach to image quality is fairly uniform, i.e., generate human imperceptible defenses by minimizing a perception heuristic like \acrfull{lpips} \cite{zhang_unreasonable_2018}.
Intuitively, an imperceptible obfuscation will retain image utility and thus be acceptable to the user.
Yet even with mainstream interest \cite{hill_this_2020}, few have adopted these systems.
More recent work interrogated the imperceptibility hypothesis finding that even when imperceptibility was achieved, it was undesirable to users \cite{logas_subversive_2024}.
Here, we build on prior literature to explore the viability and acceptability of ``intentionally perceptible'' \acrshort{afr} defenses that are designed to be aesthetically pleasing.

% However, to date, no prior work interrogated why the protected outputs of these tools are unaccepted by the average user.
% In Chapter \ref{ch:saia}, we fill this gap by evaluating user-facing \acrshort{aml} obfuscations from a human-centered perspective and propose a psychometric scale to ease the integration of human factors into adversarial obfuscation development.

\section{AuraMask: An Extensible Pipeline for Developing Aesthetic Anti-Facial Recognition Image Filters}
In this section, we detail \textsc{AuraMask}\footnote{\url{https://gitlab.com/raccs-lab/auramask-library}} --- a flexible and extensible toolkit for generating aesthetic \acrshort{afr} photo filters (``defenses'') that we use to emulate popular social media one-click image filters (\emph{RQ1}).

\begin{figure}[ht]
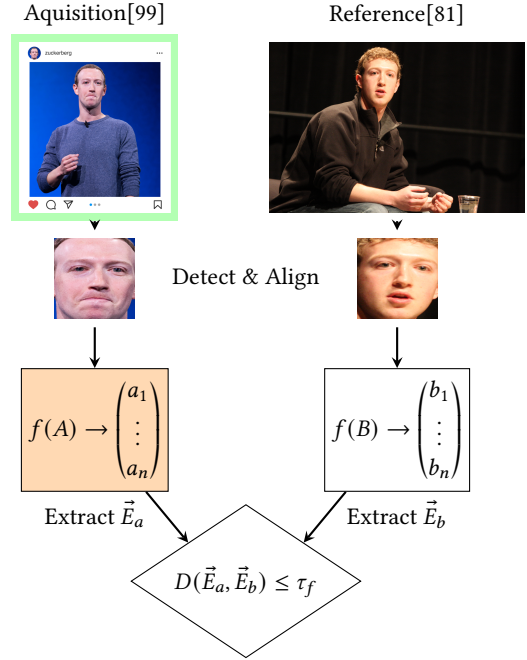

    \centering 
    \include{figs/fr-pipeline/recognitionpipeline}
    \caption{Face verification task with obfuscation opportunity in green and obfuscation target highlighted in orange.}
    \label{fig:fr-pipeline}
    \Description[Flowchart illustrating the process of facial recognition or verification, showing steps from input images through feature extraction and final distance comparison.]{}
\end{figure}

\subsection{Threat Model}
Typically, in \acrfull{aml} research, those who deploy algorithmic inference are considered ``defenders'' and those who sabotage algorithmic inference are considered ``attackers.''
We flip this framing to adopt a Subversive AI perspective \cite{das_subversive_2020} in which the inference system --- i.e., a facial recognition system --- is the ``attacker'' and the individual aiming to evade recognition is the ``defender.''

In this formulation, an attacker has (1) no prior knowledge of the victim's identity, (2) a reference image of the victim, and (3) an automated facial recognition system that matches the reference image with images connected to personal information (i.e., social media).

\textsc{AuraMask} defenses may evade one or more ``white box'', open weight face recognition models $f \in \mathbf{F}$.
However, an ideal defense should be effective against unseen models --- i.e., they should transfer to ``black box'' models.
Thus, in \cref{sec:eval}, we leave one model out of training to assess the transferability of \acrshort{afr} defenses produced with \textsc{AuraMask}.

% \textbf{Knowledge of Victim Model:} This defense is applicable in three cases. 
% First, in the case that the attacking model is known, and the defender has full access to it, then this model can be used as a defense target.
% Second, in the case that the attacking model is not known, the defender may choose an ensemble of known attackers to optimize the chance of an effective defense.

% The deployment of commercial facial recognition systems by state actors poses a significant threat to individual privacy and anonymity, enabling unprecedented levels of surveillance.
% Several systems utilize reference images -- such as those captured during protests or from CCTV footage -- to identify individuals by matching their faces to \acrshort{osn} profiles, as demonstrated by technologies like ClearView AI \cite{mac_clearviews_2020}.
% This initial identification not only exposes the victim but also their associates to further privacy invasions through additional intelligence operations, including image, human, and signals intelligence \cite{lemieux_intelligence_2024}.

% Victims of such surveillance often find themselves at a considerable disadvantage, as they lack the resources and capabilities to entirely counteract these sophisticated methods.
% Despite the challenges posed by state-sponsored surveillance, proactive obfuscation can help mitigate the risks associated with these invasive technologies.

\subsection{Defense Goal}
In common parlance, ``facial recognition'' is a catch-all term for the full recognition pipeline, its component parts -- i.e., detection, alignment, feature extraction -- or specific uses of its output -- e.g., emotion recognition, gender recognition, face verification.
We design \textsc{AuraMask} to target the \emph{feature extraction} component of the recognition pipeline, disrupting \emph{face verification}.
The feature extraction step accepts an aligned face as input ($f(x)$) and generates $N$-dimensional vectors, or \emph{face embeddings} ($\vec{E}$), as outputs.
This step is crucial to face verification as face identities are determined by the distance between embeddings.
Embeddings within a tuned distance threshold ($\tau_{f}$) are considered to belong to the same person.
In modern systems, $f$ is a \acrshort{ml} model \cite{deng_arcface_2019, cao_vggface2_2018, schroff_facenet_2015, parkhi_deep_2015} trained to minimize the distance between an identity; while retaining separability amongst individuals.
Most modern models are trained to optimize for cosine distance $D_C$:
\begin{equation}
    D_C(\vec{E_a}, \vec{E_b})= 1 - S_C(\vec{E_a}, \vec{E_b}) = 1 - \frac{\vec{E_a} \cdot \vec{E_b}}{||\vec{E_a}||\times||\vec{E_b}||}
    \label{eq:cosine-distance}
\end{equation}
Where $a$ and $b$ are images of a face, $\vec{E_{a|b}}$ are face embeddings generated from a feature extraction model $f(x)$, and $S_C(\vec{E_a}, \vec{E_b})$ denotes cosine similarity.
% These embeddings are essential for \emph{face verification}: one's face should produce a similar embedding vector across all images.
% While face embeddings have been used for many tasks -- including some that recall the days of \emph{phrenology}\footnote{I do not cite this ``research'' claiming to detect traits such as homosexuality or criminality as citations lend some amount of credibility.} -- the focus of this work is face \emph{verification} (see Figure \ref{fig:fr-pipeline}), the method used to de-anonymize online presence.
% In the face verification task, two faces are compared to determine whether they belong to the same person.
% Models trained for this purpose are optimized to generate face embeddings that effectively cluster by identity.

Prior work has shown \acrshort{aml} attacks are effective in disrupting the feature extraction step and undermining face verification accuracy \cite{shan_fawkes_2020, cherepanova_lowkey_2021, chandrasekaran_faceoff_2021, hussain_reface_2023}.
We adopt this approach in the design of \textsc{AuraMask} to sabotage an extraction model $f$, leveraging the point of control that users still have -- altering images before posting online.
Moreover, we build on prior \acrshort{afr} approaches by encouraging exploration outside ``imperceptibility'' -- i.e., minimal noticeable difference between $x$ (the original image) and $\hat{x}$ (the perturbed image) -- with a modular pipeline design for experimentation with aesthetic heuristics -- e.g., emulation of popular image filters.
In this work, we assume face verification is performed by comparing one altered and one unaltered image.
This assumption would hold if, an adversarial actor took a photo of someone in the physical world (unaltered) and tried to match it to their social media profile picture (altered).

\subsection{Problem Formulation}
The optimization problem in the \textsc{AuraMask} pipeline is twofold:
\begin{enumerate}
    \item The face in the output image cannot be algorithmically recognized as the same face in an unaltered image.
    \item The output image aligns with some heuristic for image aesthetics.
\end{enumerate}

Integrating aesthetics as a parallel objective may, at first, seem disconnected or indeed orthogonal to the protection objective.
Note, however, that prior approaches \cite{cherepanova_lowkey_2021, shan_fawkes_2020, hussain_reface_2023} optimize for both adversarial effect and proximal perceptual metrics to generate ``imperceptible'' defenses.
These optimization heuristics -- i.e., \acrfull{lpips} amongst others -- were critiqued for inadequately aligning with human perception \cite{sen_popular_2020} and how the resulting defenses were perceived by end users \cite{logas_subversive_2024}.

Thus, we re-contextualize and extend the defense objective with a \acrfull{mtl} approach \cite{yu_multitask_2014, zhao_multitask_2023}.
Within \acrshort{mtl}, a network is concurrently optimized for two or more distinct but conceptually similar tasks.
Here, the two tasks were to sabotage the facial recognition pipeline and to generate outputs that align with an aesthetic concept.
These two tasks are promising for a \acrshort{mtl} approach as they both accomplish this goal through altering image pixels.
To investigate this theoretical synergy, we designed the \textsc{AuraMask} pipeline to generate image filters that defend against facial recognition -- i.e., $\max D_{C}$ -- while concurrently optimizing for one or more aesthetic heuristics -- i.e., $\mathcal{L}_{AES}$.
% First, the \acrshort{atn} is trained to generate outputs $g(x) \rightarrow \hat{x}$ such that $D_{C}(\vec{E_{\hat{x}}},\vec{E_{x}}) \geq \tau_{f}$ for a targeted model $f$.
% Second, the generated output $\hat{x}$ aligns with an aesthetic target -- e.g., emulate a popular one-click image filter $i(x) \rightarrow x_{i} \approx g(x) \rightarrow \hat{x}_{i}$.

\subsubsection{Objective 1: Sabotaging Face Embeddings}
\begin{figure}[ht]
    \centering
    \begin{tikzpicture}[domain=0:2,xscale=3.1,yscale=1.75]
        \fill[Green!20] (0,-1.05) grid (0.6,1.05) rectangle (0,-1) node [above] at (0.3,1) {\small\textcolor{Green}{Face Verified}\medskip};
        \draw[thin,step=0.1,color=Gray!40] (0,-1.05) grid (2.05,1.05);

        \foreach \i in {0.5,1.0,1.5} {
            \draw [very thin,Gray] (\i,-1) -- (\i,1)  node [below] at (\i,-1) {$\i$};
        }
        \foreach \i in {-0.5,0.5} {
            \draw [very thin,Gray] (0,\i) -- (2,\i) node [left] at (0,\i) {$\i$};
        }

        \draw[->] (-0.1,0) -- (2.1,0) node[right] {$D_{C}$};
        \draw[->] (0,-1) -- (0,1.2) node[above] {$\mathcal{L}$};

        \draw[dotted,Green, line width=2pt] (0.6,-1) -- (0.6,1.2) node[above] {$\tau_{f}$};
      
        \draw[color=red, line width=1pt]    plot (\x,{1-\x})             node[right] {\acrshort{fe}};
        \draw[color=blue, line width=1pt]   plot (\x,{abs(1 - \x)})    node[right] {\acrshort{fea}};
        \draw[color=orange, line width=1.2pt] plot (\x,{abs(1 - \x) - (1-0.6)}) node[right] {\acrshort{feat}};
    \end{tikzpicture}
    \caption{Plot of face embedding losses with respect to cosine distance, highlighting where faces are considered validated for $\tau=0.6$}
    \label{fig:femb-losses}
\end{figure}
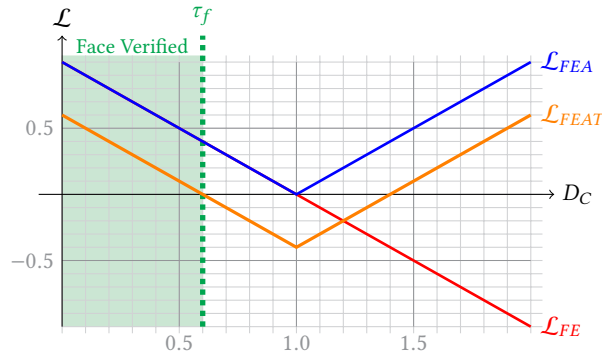
To increase the distance between $x$ and $\hat{x}$, prior methods adopt the \acrfull{fe} loss \cite{hussain_reface_2023} which optimizes for the maximum cosine distance or minimum cosine similarity -- i.e., opposite cosine direction.
\begin{equation}
    \arg \min_{\hat{x}} S_C(f(x), f(\hat{x}))
    \label{eq:fitness-function-fe}
\end{equation}
where $f$ is the targeted face embedding model, $x$ is the unaltered image, $\hat{x}$ is the adversarially perturbed image generated by $g$, and $S_C$ is the cosine similarity \ref{eq:cosine-distance}.
The downside of \acrshort{fe} is its greedy nature, encouraging optimization well beyond orthogonality which is generally unnecessary to sabotage face verification.

Thus, to address over-optimization which could add unnecessary perturbations, we propose two alterations to \acrshort{fe}.
First, the \acrfull{fea} loss seeks to limit over-optimization by penalizing distances beyond orthogonality by taking the absolute value of \acrshort{fe}:
\begin{equation}
    \arg \min_{\hat{x}} |S_{C}(f(x), f(\hat{x}))|
    \label{eq:fitness-function-fea}
\end{equation}
However, this loss does not directly target the verification task nor is it particularly descriptive -- e.g., a loss of 0.5 doesn't meaningfully convey effectiveness against face verification.

Our second alteration incorporates the tuned distance threshold ($\tau_f$) for a given model -- under which embeddings are considered to belong to the same person -- into the \acrshort{fea} loss resulting in the \acrfull{feat} loss.
\begin{equation}
    \arg \min_{\hat{x}} |S_{C}(f(x), f(\hat{x}))| - \tau_f
    \label{eq:fitness-function-feat}
\end{equation}
This loss both targets the decision boundary for face verification while constraining against over-optimization.
\acrshort{feat} also improves upon interpret-ability over \acrshort{fea} as a negative value demonstrates effectiveness against face verification.

\subsubsection{Objective 2: Integrating Aesthetics}
% TODO: Introduce integrating aesthetics from a more general perspective and then talk about how we integrate instagram filters for training
Unlike the prior objective, the aesthetic objective -- $\mathcal{L}_{AES}$ -- has no singularly agreed upon optimum given that much of ``aesthetic'' evaluation is subjective.
Thus, instead of providing a prescriptive aesthetic heuristic for \acrshort{afr} defenses, we designed \textsc{AuraMask} to interchangeably accept any aesthetic heuristic by leveraging the backend-agnostic Keras 3 library.
This design eases the use of both existing and novel ``aesthetic'' heuristics when implemented with the \texttt{keras.Loss} class, which accepts batched image input and performs differentiable computations to arrive at a numeric output.
Through this approach, novel \acrshort{afr} defenses may be integrated into a range of image aesthetic enhancement methods -- e.g., style transfer \cite{shih_style_2014}, learned quality assessments \cite{talebi_nima_2018}, or computational photography methods \cite{delbracio_mobile_2021,machado_computing_1998}.
Note, however, that learned image heuristics ($\mathcal{L}_{LRND}$) may share latent embeddings with facial recognition models and may therefore be disrupted by improved adversarial effect -- e.g., \acrshort{fe}, \acrshort{fea}, or \acrshort{feat}.
Thus, we recommend inclusion of a closed-form loss ($\mathcal{L}_{CLSD}$) -- i.e., \acrshort{mse}, \acrshort{mae}, \acrshort{ssim} -- when using a learned heuristic.
A simple $\mathcal{L}_{AES}$ may be defined as:
\begin{equation}
    \arg \min_{\hat{x}} \mathcal{L}_{LRND}(x, \hat{x}) + \mathcal{L}_{CLSD}(x, \hat{x})
\end{equation}

With the two objectives (\acrshort{feat} and $\mathcal{L}_{AES}$) we thus seek to optimize the defense ($g$) such that:
\begin{equation}
    \forall_{x \in \mathbf{X}} \arg \min_{x} \mathcal{L}_{FEAT}(x, g(x)) + \mathcal{L}_{AES}(x, g(x))
    \label{eq:nofilt-fitness-function}
\end{equation}

\subsection{Technical Infrastructure and Definitions}
Many prior \acrshort{aml}-based obfuscations use iterative methods to generate effective outputs through multiple forward and backward passes over target models (e.g., \acrfull{pgd} \cite{madry_deep_2019} or the \acrfull{fgsm} \cite{akhtar_threat_2018}). 
We designed the \textsc{AuraMask} pipeline to instead take advantage of the \acrfull{atn} \cite{baluja_adversarial_2017, hussain_reface_2023}.
Unlike \acrshort{fgsm} and \acrshort{pgd}, the \acrshort{atn} learns to predict an effective adversarial perturbation.
As such, it can apply a defense with only a single forward pass and does not need to access the targeted model(s) when applying this pass.
This is because the \acrfull{atn} is trained to transform an input into an adversarial example against one or more target networks.
Such networks can be targeted or untargeted, trained for white-box or black-box contexts.
Formally, an \acrshort{atn} \cite{baluja_adversarial_2017} is defined as:
\begin{equation}
    g_{\mathbf{F}}(x) \colon x \in X \rightarrow \hat{x}
    \label{eq:atn-formal}
\end{equation}
where $\mathbf{F}$ is the set of target models and $x \approx \hat{x}$ but $f(x) \neq f(\hat{x}) \forall f \in \mathbf{F}$.
At inference time, $g_{\mathbf{F}}$ may be run on any input $x$ without access to $f \in \mathbf{F}$ or further gradient computations.
These properties make an \acrshort{atn} much faster to use than even single iteration of \acrshort{fgsm}.
However, such an approach introduces the complexity of training data, model architecture, and training hyperparameters.

\subsubsection{Training Data}
We designed \textsc{AuraMask} with in-built support for three face-based training datasets and interfaces for easy integration of new datasets.\\

\noindent\textit{\acrfull{fdf}} \cite{hukkelas_deepprivacy_2019} consists of 1.5 million faces collected from the Flickr image sharing site and is notable for labeling images with face bounding boxes and including copyright licenses.
Images in this dataset have a diverse distribution in pose, age, ethnicity, occlusions, face paint, and image background.\\

\noindent\textit{\acrfull{lfw}} \cite{huang_labeled_2008} consists of 13.2k unique images featuring people from a diverse population and captured under various conditions and poses.
A specialized subset of 2.2k image pairs is used to benchmark face recognition models.
Each image pair is labeled as the same person (1) or different people (0), balanced with 1.1k for each class.\\

\noindent\textit{VGGFace2} \cite{cao_vggface2_2018} consists of 3.3 million images and which represent 10,000 individuals collected from Google Image Search and manually labeled for identity.
As with \acrshort{fdf}, this dataset claims a diversity in subject, pose, and quality.

\subsubsection{Model Architecture and Hyperparameters}
The objectives we formalize above require a model architecture that supports ingesting an image ($x$) and generating an altered version of that image ($\hat{x}$) -- i.e., an image to image model.
The U-Net, with its symmetric encoder and decoder, has emerged as a powerful and dynamic architecture for image to image tasks from segmentation \cite{ronneberger_unet_2015} to resolution enhancement \cite{rombach_highresolution_2022}.
Thus, we implemented five UNet-based architectures in \textsc{AuraMask} to use as a basis for creating novel \acrshort{atn} defenses\footnote{Modernized from \cite{sha_kerasunetcollection_2021}}: UNet \cite{ronneberger_unet_2015}, VNet \cite{milletari_vnet_2016}, R2U-Net \cite{alom_recurrent_2018}, Attention U-Net \cite{oktay_attention_2018}, ResUnet-a \cite{diakogiannis_resuneta_2020}.
In \textsc{AuraMask}, each architecture may be used with a default set of structural parameters or may be customized with a JSON file to ease hyperparameter experimentation.

\subsection{Implementation} \label{sec:poc-implementation}
We use the pipeline detailed above to create a proof-of-concept set of ``aesthetic'' \acrshort{afr} defenses to address RQ2 and RQ3.
The ``aesthetic'' heuristic we use is similarity to existing image filters available on the Instagram photo-sharing \acrshort{osn} given prior work found ``that users like to apply [Instagram] filters on their photos even though it is a time-consuming process and requires spending more effort'' \cite{bakhshi_why_2015}.
In total, we generated 80 Instagram filter-like defenses: 40 \acrfull{singletarget} -- trained against the ArcFace \cite{deng_arcface_2019} embedding model -- and 40 \acrfull{ensembletarget} -- trained against both ArcFace and VGGFace2 \cite{cao_vggface2_2018} embedding models.

\subsubsection{$\mathcal{L}_{AES}$: Emulating Aesthetic Image Filters}
In the context of the two objectives described before, \emph{our} second objective for the \textsc{AuraMask} defenses we created was to introduce alterations similar to an image filter known to be aesthetically pleasing.
Formally, for a given graphical filter ($i \in \mathbb{I}$) we seek to minimize the error between the generated output ($g(x) \rightarrow \hat{x}$) and the output of a graphical filter ($i(x) \rightarrow \tilde{x}$) such that $\hat{x} \approxeq \tilde{x}$.
To date, no dataset exists to meet both these tasks; thus, we leveraged the \texttt{pilgram2}\footnote{\url{https://github.com/mgineer85/pilgram2}} library which programmatically applies a given Instagram filter ($i \in \mathbb{I}$).
This approach has the advantage of being applicable to any image dataset since the target outputs ($\tilde{x}$) are generated at training time.

We trained both \acrshort{singletarget} and \acrshort{ensembletarget} defenses by minimizing the difference between $\hat{x}$ and $\tilde{x}$ as measured by a classically-trained, full-reference perceptual loss heuristic and a closed-form heuristic.
The learned heuristic, $\mathcal{L}_{TopIQ}$ \cite{chen_topiq_2023}, accepts two images and generates a score between 0 and 1 representing the perceptual similarity between the images.
However, as mentioned previously, learned heuristics can be negatively impacted by \acrshort{afr} optimizations\footnote{See Figure \ref{fig:topiq-only}}.
Thus, we include a closed-form heuristic to mitigate interference on the learned heuristic.
In our \acrshort{singletarget}s, we found $\mathcal{L}_{MSE}$ -- mean squared error based on pixels -- to sufficiently mitigate the effect of \acrshort{afr} optimizations on $\mathcal{L}_{TopIQ}$.
The \acrshort{ensembletarget}s, however, proved more susceptible to inadvertent heuristic interference which $\mathcal{L}_{MSE}$ was unable to mitigate, so we adopted $\mathcal{L}_{SSIM}$ -- a metric which measures image similarity based on luminance, contrast, and structure \cite{venkataramanan_hitchhikers_2021}.
We formalize $\mathcal{L}_{AES}$ in this work as:

\begin{equation}
    \acrshort{singletarget} \mathcal{L}_{AES}: \frac{\mathcal{L}_{TopIQ} + \mathcal{L}_{MSE}}{2}
    \label{eq:st-aes-fitness}
\end{equation}

\begin{equation}
    \acrshort{ensembletarget} \mathcal{L}_{AES}: \frac{\mathcal{L}_{TopIQ} + \mathcal{L}_{SSIM}}{2}
    \label{eq:et-aes-fitness}
\end{equation}

\subsubsection{Model Architecture}
Our proof-of-concept defenses were trained on the VNet architecture, modified for 2D \cite{sha_kerasunetcollection_2021}.
We took this approach given the architectural implications raised by Hussain et al. \cite{hussain_reface_2023} -- demonstrating that the standard convolutions of a UNet were insufficient to defend against facial recognition but a stack of convolutions with recurrent connections performed well.
Thus, we use five recurrent convolutions for the encoding and decoding steps starting with a width of 64 channels and doubling at each step except the last -- i.e. [64, 128, 256, 512, 512].
The number of stacked encoding convolutions is mirrored in the decoding path, starting with the initial stack of one and increasing to a maximum of three.
In addition, we confirmed that using pooling and unpooling layers for down and up-sampling were detrimental to adversarial defense effectiveness.
% \lstinputlisting[language=json, caption=Proof of concept architecture configuration, label=lst:config]{files/arcface_vnet.json}

\subsubsection{Training Procedure}
We trained each of our 80 proof-of-concept \acrshort{afr} defenses on an NVIDIA H100 GPU over 500 epochs which performed 50 training steps of 64 256x256 images.
VNet weights were optimized using AdamW \cite{loshchilov_decoupled_2019} with a learning rate of $1e-4$.

Our training dataset, \acrfull{fdf} \cite{hukkelas_deepprivacy_2019}, consists of 1.5 million face images collected from the Flickr image sharing service with a diverse distribution of age, pose, ethnicity, occlusion, face paint, and background.
In training, we use a subset of images with a size greater than or equal to 256x256 -- about 4\% of the total dataset -- of which we take a 90/10 split resulting in 217k training and 24k validation images.
We chose this subset to avoid interpolation artifacts that upscaling could introduce.

When training, images were preprocessed by resizing to 256x256, center cropping to 224x224, and applying FancyPCA -- a technique to capture ``an important property of natural images, i.e., that identity is invariant to changes in the intensity and color of the illumination'' \cite{krizhevsky_imagenet_2012}.
We also randomly apply geometric augmentations -- i.e., either vertical or horizontal flip -- at a rate of $0.5$ and non-geometric augmentations -- i.e., guassian blurring, gaussian noise, or image sharpening -- at a rate of $0.2$ to the training data.
Finally, we generated the target output by applying a given \texttt{pilgram2} filter to the training data.

% \subsubsection{User Acceptance: Quality Measures}
% Beyond the user study, we also compute the various heuristics for image quality and perceptibility \cite{sen_popular_2020} as proximal measures for aesthetic acceptability: \acrfull{mse}, \acrfull{mae}, \acrfull{ssim} \cite{venkataramanan_hitchhikers_2021}, \acrshort{lpips} \cite{zhang_unreasonable_2018}.
% In addition, we include a more recent perceptual image quality measure: TopIQ \cite{chen_topiq_2023}.
% % This experiment provides an accounting for quality and heuristics across the novel and prior obfuscation.

\section{Technical Evaluation} \label{sec:eval}
To address \emph{RQ2}, we calculated how well our proof-of-concept aesthetic \acrshort{afr} defenses compared to existing defenses (i.e., Fawkes and LowKey) at evading facial recognition across a range of scenarios.

\subsection{Evaluation Configuration}

\subsubsection{Datasets}
We use the test split of the \acrshort{fdf} dataset and the ``pairs'' configuration of the \acrfull{lfw} \cite{learned-miller_labeled_2016} dataset.

\noindent \textbf{\acrfull{fdf}} We evaluated adversarial and quality performance with the testing split which has 6.53k unique images.
When testing, images were resized to 256x256 and center cropped to 224x224 with no other preprocessing alterations applied.

\noindent \textbf{\acrfull{lfw}} Consists of 13.2k unique images featuring people from a diverse population and captured under various conditions and poses.
Within \acrshort{lfw}, a specialized subset of 2.2k image pairs is often used as a benchmark to compare novel face recognition models \cite{mohapatra_comparative_2025}.
Each image pair is labeled as either the same person (1) or different people (0), balanced with 1.1k pairs for each class.
We use this subset to measure the impact each defense has on the recall for the 1.1k pairs that represent true positive in the face verification task.

\subsubsection{Baseline Controls}
We compared our ``aesthetic'' defenses against two prior \acrshort{pgd}-based \acrshort{afr} defenses -- Fawkes \cite{shan_fawkes_2020} and LowKey \cite{cherepanova_lowkey_2021} -- which were optimized for imperceptibility.
While we do not formally report on obfuscation generation time against these baseline defenses, the \acrshort{atn} method that \textsc{AuraMask}-based defenses use is significantly faster than \acrshort{pgd}-based approaches (1-3s versus minutes in some cases) \cite{hussain_reface_2023}.

\subsubsection{Target Facial Recognition Models}
We evaluated adversarial effectiveness against pretrained implementations of four facial recognition models from the DeepFace Python repository: ArcFace \cite{deng_arcface_2019}, VGGFace2 \cite{cao_vggface2_2018}, and Facenet \cite{schroff_facenet_2015} \footnote{\url{https://github.com/serengil/deepface}}\cite{serengil_benchmark_2024, serengil_lightface_2020}.
% We chose four facial recognition models to compare against to be comprehensive, but doing so presented a challenge for the user acceptance evaluation as the filters AuraMask produced to target each of these facial recognition models varied.
% For simplicity, therefore, we conducted the user acceptance evaluation only on the filters trained to target ArcFace.
% This evaluation still allows to compare relative user acceptance between AuraMask filters and the control conditions we defined above, even if the absolute user acceptance of AuraMask filters may vary when targeting different facial recognition models.

% Of these models, the \acrshort{atn}s evaluated with \acrshort{saia8} are only trained to target ArcFace, however all other evaluations are performed on models trained on ArcFace, VGGFace, Facenet512, and an ensemble of ArcFace and VGGFace.
% We chose to perform \acrshort{saia8} evaluations on ArcFace trained outputs as the research team agreed the obfuscations against the other models were much more extreme thus we leave acceptability tuning of these to future work.
\subsection{Adversarial Effectiveness: Evaluation Metrics}
We measured adversarial effectiveness of the defenses we tested (RQ2) using both the distance divergence and face verification recall metrics.
% , and user acceptance (RQ3) using both metrics from the computational aesthetics literature and a validated scale of user acceptance of privacy-enhancing image perturbations from the CSCW literature --- the Subversive AI Acceptance Scale (\acrshort{saia8}) \cite{logas_subversive_2024}.

\subsubsection{Distance Divergence}
Distance divergence measures how successful each defense is in moving the cosine distance of an image beyond the verification threshold for a given face embedding.
We quantify distance divergence as the ratio of image pairs ($x$, $\hat{x}$) in the test split of the \acrshort{fdf} dataset that fall within the verification threshold for a given face embedding model ($f$).
% In essence, we perform face validation for pairs of unaltered and obfuscated images.
This metric represents the worst-case difficulty for face verification defense, where the image using to query is present in unaltered form in the facial recognition database.
% This data does demonstrate the adversarial effectiveness of an obfuscation in the worst-case --- e.g., $x$ vs $\hat{x}$.

%Note, that this metric is \emph{not} clearly aligned with real-world face verification threat.
%However, this data does demonstrate the adversarial effectiveness of an obfuscation in the worst-case --- e.g., $x$ vs $\hat{x}$.

\subsubsection{Face Verification Recall} \label{sec:face-ver-recall}
Face verification recall measures how well each defense sabotages face verification recall performance for pairs of unique images using the ``positive pairs'' subset of the \acrshort{lfw} benchmark.
In evaluation, we quantify face verification recall as the ratio of true positives -- i.e., cosine distance accurately falls within a given threshold ($\tau_{f}$) -- over the total number of positive pairs in the test set \cite{serengil_benchmark_2024,serengil_lightface_2020}.
As face verification works best on inputs cropped to a face, we used an off-the-shelf implementation of MTCNN \cite{zhang_joint_2016} to detect and crop to faces for both images in the pair.
We only consider true positive pairs (i.e., recall rate), as our primary aim was to increase false negatives and not false positives in face verification.
This metric serves as a comparison point between prior defenses and our ``aesthetic'' defenses in reducing the effectiveness of face recognition pipelines.

In addition, we evaluated the recall rate for face verification when both images of a pair are obfuscated.
While this case is implicitly excluded by our threat model (which assumed that the attacker has only an unaltered image of the subject), we performed this evaluation to understand the effectiveness of the tested \acrshort{afr} defenses in the more pessimistic case wherein an attacker also has a different obfuscated image of the subject.
% consider the impact of repeated use of the proposed obfuscations -- i.e., if regular use results in a new, but equally identifiable embedding.
We performed two sets of evaluation with eight of the \acrshort{afr} \textsc{AuraMask}-based defenses -- four \emph{\acrfull{singletarget}} and four \emph{\acrfull{ensembletarget}}.
% The results on this metric provide a better understanding of how repeated, casual use of these defenses could impact the overall protective effect.

For brevity, we present the adversarial and quality measures for four of the \acrshort{singletarget} and \acrshort{ensembletarget} defenses: i.e., Dogpatch, Nashville, Sutro, and Moon\footnote{Full table of results in Appendix~\ref{tab:full-accuracy}}.

\subsection{Results}
\subsubsection{Distance Divergence}

\begin{table}[ht]
    \centering
    \begin{tabular}{lcccc}
        % \multicolumn{5}{c}{\textbf{Recall Rate with \acrshort{fdf} Testing Data}} \\
        \toprule
        \multicolumn{2}{c}{}                                                                              & ArcFace \cite{deng_arcface_2019} & VGGFace \cite{cao_vggface2_2018}  & Facenet \cite{schroff_facenet_2015}    \\ \cmidrule(lr){3-5}
        \multicolumn{2}{c}{Baseline}                                                                      & $1.000$                          & $1.000$                           & $1.000$                                \\ \cmidrule(lr){3-5}
        \multicolumn{2}{c}{LowKey \cite{cherepanova_lowkey_2021}}                                         & $0.974$                          & $0.970$                           & $0.470$                                \\ \cmidrule(lr){3-5}
        \multirow{3}{*}{Fawkes \cite{shan_fawkes_2020}}             & \textit{L}                          & $0.993$                          & $1.000$                           & $0.888$                                \\
                                                                    & \textit{M}                          & $0.921$                          & $0.998$                           & $0.643$                                \\
                                                                    & \textit{H}                          & $0.864$                          & $0.997$                           & $0.519$                                \\ \cmidrule(lr){3-5}
        \multirow{2}{*}{Dogpatch}                                   & \textit{\acrshort{singletarget}}    & $0.006$                          & $0.996$                           & $0.956$                                \\
                                                                    & \textit{\acrshort{ensembletarget}}  & $0.087$                          & $0.200$                           & $0.339$                                \\ \cmidrule(lr){3-5}
        \multirow{2}{*}{Moon}                                       & \textit{\acrshort{singletarget}}    & $0.008$                          & $0.996$                           & $0.874$                                \\
                                                                    & \textit{\acrshort{ensembletarget}}  & $0.011$                          & $0.077$                           & $0.096$                                \\ \cmidrule(lr){3-5}
        \multirow{2}{*}{Nashville}                                  & \textit{\acrshort{singletarget}}    & $0.015$                          & $1.000$                           & $0.965$                                \\
                                                                    & \textit{\acrshort{ensembletarget}}  & $0.052$                          & $0.107$                           & $0.244$                                \\ \cmidrule(lr){3-5}
        \multirow{2}{*}{Sutro}                                      & \textit{\acrshort{singletarget}}    & $0.006$                          & $1.000$                           & $0.969$                                \\
                                                                    & \textit{\acrshort{ensembletarget}}  & $0.024$                          & $0.128$                           & $0.171$                                \\ \bottomrule
    \end{tabular}
    \caption{Face verification recall rate with \acrshort{fdf} testing subset -- verifying $x$ and $\hat{x}$ pairs.}
    \label{tab:fdf-accuracy}
\end{table}

In Table~\ref{tab:fdf-accuracy}, we report the face verification accuracy on the test split of \acrshort{fdf} where $x$ and $\hat{x}$ are pairs.
We can see from this data that the \acrshort{singletarget}s outperform the prior methods and the \acrshort{ensembletarget}s when applied to ArcFace -- the model it was trained against.
However, it is also evident that this performance does not transfer well to unseen, ``black box'' models -- i.e. Facenet and VGGFace2.
\acrshort{ensembletarget}s, on the other hand, outperform prior methods when used against either ArcFace or VGGFace2 -- the models it was trained against -- as well as the ``black box'' holdout Facenet.
Of note is the Moon \acrshort{ensembletarget} which performed best overall -- reducing the accuracy across models by an average of $93.9\%$.
These results align with the findings of \cite{hussain_reface_2023} which also demonstrated lower attack transfer than that of \acrshort{pgd} methods when trained against a single embedding model.
Across all defenses under test, VGGFace2 \cite{cao_vggface2_2018} proved to be the most difficult embedding model to sabotage.

\subsubsection{Face Verification Recall}

\begin{figure}[h]
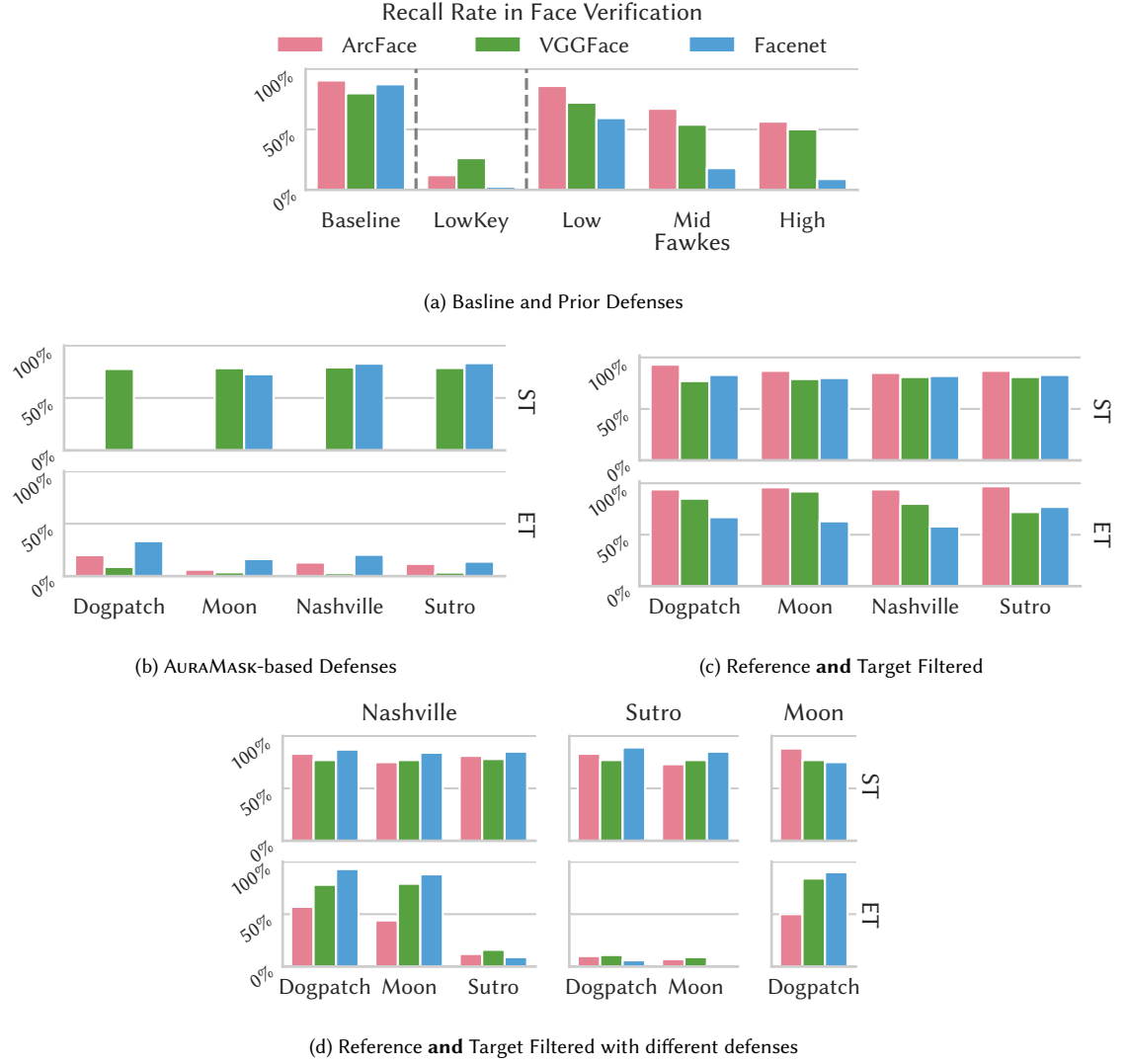

    \centering
    \begin{subfigure}{\textwidth}
        \centering
        \input{figs/baseline-performance.pgf}
        \caption{Basline and Prior Defenses}
        \label{fig:baseline}
        \Description[]{Baseline and Prior Defenses: Displays recall under varying defenses (e.g., LowKey, Mid Fawks), establishing a benchmark for model performance before novel defenses are applied.}
    \end{subfigure}
    \hfill
    \begin{subfigure}{0.49\textwidth}
        \input{figs/defense-performance.pgf}
        \caption{\textsc{AuraMask}-based Defenses}
        \label{fig:lfw-precision-arcface}
        \Description[]{AURAMASK-based Defenses: Shows Single Target (ST) and Ensemble Target (ET) recall rates for four different defenses (Dogpatch, Moon, Nashville, Sutro).}
    \end{subfigure}
    \hfill
    \begin{subfigure}{0.49\textwidth}
        \input{figs/obfuscation-pairs.pgf}
        \caption{Reference \textbf{and} Target Filtered}
        \label{fig:both-obfuscated}
        \Description[]{Presents Single Target and Ensemble Target recall under the filtering methodology when evaluated with pairs of obfuscated images using the same model.}
    \end{subfigure}
    \hfill
    \begin{subfigure}{\textwidth}
        \centering
        \input{figs/obfuscations-mixed.pgf}
        \caption{Reference \textbf{and} Target Filtered with different defenses}
        \label{fig:mixed-obfuscated}
        \Description[]{Presents Single Target and Ensemble Target recall under the filtering methodology when evaluated with pairs of obufscated images using different models.}
    \end{subfigure}
    \caption{\acrshort{lfw} pairwise face verification recall using ArcFace, VGGFace2, Facenet embeddings across defenses.}
    \label{fig:lfw-precision-recall}
    \Description[]{Figure 4: LFW Pairwise Face Verification Recall Across Defenses. This multi-panel graph compares the recall rates of three embedding models—ArcFace, VGGFace2, and Facenet—when performing pairwise face verification using the LFW dataset. The analysis is broken down across four types of defense scenarios}
\end{figure}

In Figure~\ref{fig:lfw-precision-recall}, we summarize the results of the comparative evaluation on the \acrlong{lfw} pairs benchmark\footnote{Full table of results in Appendix~\ref{tab:full-accuracy}}.

With respect to face verification recall --- i.e., correct positive predictions over the total number of true positives --- we can see that both the \acrshort{singletarget} and \acrshort{ensembletarget} defenses are effective in reducing recall over the baseline.
In particular, \acrshort{ensembletarget}s are as effective as LowKey and outperform Fawkes across all tested face embeddings.
The \acrshort{singletarget}s, on the other hand, are generally only effective in sabotaging verification with the targeted model (ArcFace) -- a result that mirrors that of prior work on \acrshort{atn}s \cite{hussain_reface_2023}.

Next, consider Figure~\ref{fig:both-obfuscated}, which illustrates the reduced protection when both images in a pair are obfuscated by the same defense.
This means that an attacker with access to a reference image obfuscated by the same defense could negate the protection provided.
However, in Figure~\ref{fig:mixed-obfuscated} we can see that using a mix of defenses, especially with the \acrshort{ensembletarget}, retains more protection.
% We discuss the implications of these results below.

\subsubsection{Perceptual Similarity}
\begin{figure}[ht]
    \input{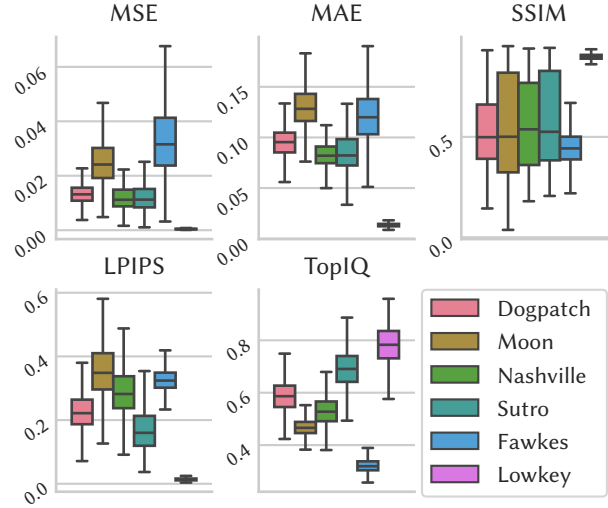}
    \caption{Perceptual similarity assessment across defenses.}
    \label{fig:quality-metrics}
    \Description[Figure showing box plots of various perceptual similarity metrics—including LPIPS, MSE, MAE, SSIM, and TopIQ—comparing their values across six different defensive techniques.]{Figure 5: Perceptual Similarity Assessment Across Defenses. This composite figure consists of five box plots, each measuring a different perceptual similarity metric—LPIPS (Learned Perceptual Image Patch Similarity), MSE (Mean Squared Error), MAE (Mean Absolute Error), SSIM (Structural Similarity Index Measure), and TopIQ. Each subplot displays the distribution of the chosen metric across six specific defensive techniques: Dogpatch, Moon, Nashville, Sutro, Fawkes, and Lowkey. The box plots show the median, quartiles, and range of values for each defense method for that specific metric. Key Observations: Error Metrics (LPIPS, MSE, MAE): These metrics typically measure deviation or error. In most cases, lower values across these charts indicate less distortion or higher fidelity in the transformed images due to the applied defense. For instance, Lowkey shows generally low distributions for LPIPS and TopIQ. Similarity Metrics (SSIM): This metric measures structural similarity, where higher values are generally better. The SSIM plot shows that most defenses yield relatively high and comparable ranges of similarity scores. The overall purpose of the figure is to quantitatively assess how well different defensive mechanisms preserve the perceptual quality and fidelity of an image while applying countermeasures.}
\end{figure}

In Figure~\ref{fig:quality-metrics}, we present the distribution of multiple commonly used perceptual similarity heuristics \cite{sen_popular_2020} -- i.e., \acrshort{mse}, \acrshort{mae}, \acrshort{ssim}, \acrshort{lpips} -- and a more recent perceptual heuristic -- TopIQ \cite{chen_topiq_2023} -- on the \acrshort{fdf} testing subset.
As expected, prior \acrshort{afr} defenses generate outputs that are more perceptually similar than our ``aesthetic'' defenses with perturbations introduced by LowKey \cite{cherepanova_lowkey_2021} measured as the least perceptible.

\section{User Evaluation}
To address RQ3, we conducted a controlled study to evaluate user acceptance of \textsc{AuraMask} defenses relative to the baseline conditions.
We collected user data across $N=630$ participants with a two-task survey wherein participants reported on the ``acceptability'' of outputs generated by a randomly selected defense using a validated scale --- the \acrfull{saia8} \cite{logas_subversive_2024}.
In the first task, participants responded to the \acrshort{saia8} scale across seven conditions \cite{logas_subversive_2024}.
In the second task, participants were asked to choose up to three from a set of fourteen \acrshort{afr} defenses --- e.g., Figure \ref{fig:obfuscation-examples} --- that they would feel comfortable using on their own photos.
% Participants were presented with an image before filtering and after filtering with one of seven conditions: four AuraMask defenses, the Fawkes defense \cite{shan_fawkes_2020}, the LowKey defense \cite{cherepanova_lowkey_2021}, or a control of no defense.
% With this data, we performed a one-way \acrfull{anova} \cite{st_analysis_1989} followed by the Tukey test \cite{tukey_comparing_1949} to confirm and explore statistical differences between reported acceptance of the presented obfuscations.
Our analytical approach included a mix of methods including one-way \acrfull{anova} \cite{st_analysis_1989}, the Tukey test \cite{tukey_comparing_1949}, and Plackett-Luce \cite{turner_plackettluce_2025} to evaluate the relationship in user responses amongst the control, baseline defenses, and our aesthetic defenses.

\subsection{Conditions}
In this study, we evaluated a subset of the 80 defenses we generated with \textsc{AuraMask} as our focus was on measuring the difference in user acceptability between intentionally perceptible \acrshort{afr} defenses and prior approaches, not between the \textsc{AuraMask} defenses.
Accordingly, we chose four \textsc{AuraMask} defenses for the first user task (rating acceptability on the \acrshort{saia8} scale) and twelve for the second user task (top 3 preference selection).
The subsets were chosen, in part, due to the diversity of effects they represent; for example, the four defenses used for the first task: Moon is black-and-white, Nashville and Dogpatch vary image warmth, and Sutro introduces shadows at the image border along with color variation.
Finally, we only tested the \acrfull{singletarget} variant in our user study --- trained against ArcFace \cite{deng_arcface_2019} --- which had lower overall protection (see \ref{sec:face-ver-recall}).
% We leave it to future work to more comprehensively evaluate user preference against a broader set of variations in aesthetic style and target model.

\subsubsection{Task 1: Self-Reported Acceptability}
In the first task, we randomly assigned each participant one of seven conditions --- i.e., Control (no change), Fawkes \cite{shan_fawkes_2020}, LowKey \cite{cherepanova_lowkey_2021}, Dogpatch, Moon, Nashville, or Sutro.

\subsubsection{Task 2: Multi-Choice Grid Selection}
In the second task, each participant was presented with fourteen defenses in a grid -- 12 \acrshort{singletarget} \textsc{AuraMask}-based defenses (see Figure~\ref{fig:obfuscation-examples}) along with Fawkes and LowKey.

\subsection{Image Preparation}
% We chose to filter images before the survey instead of on demand filtering of personal images as \acrshort{pgd}-based obfuscation methods -- e.g., Fawkes\cite{shan_fawkes_2020} and LowKey\cite{cherepanova_lowkey_2021} -- take up to 15 minutes to complete.
% While we expect that \acrshort{atn}-style obfuscations like those that are produced with the AuraMask pipeline will be more user acceptable in terms of speed \cite{hussain_reface_2023}, our focus in this study was on the comparative acceptability between perceptible-but-aesthetic versus ``imperceptible'' \acrshort{afr} filter outputs.
We pre-applied defenses to a randomly sampled subset of 1K photos from the \acrshort{fdf} test split across all aforementioned conditions.

\begin{figure}[h]
    \centering
    \includegraphics[width=0.8\linewidth]{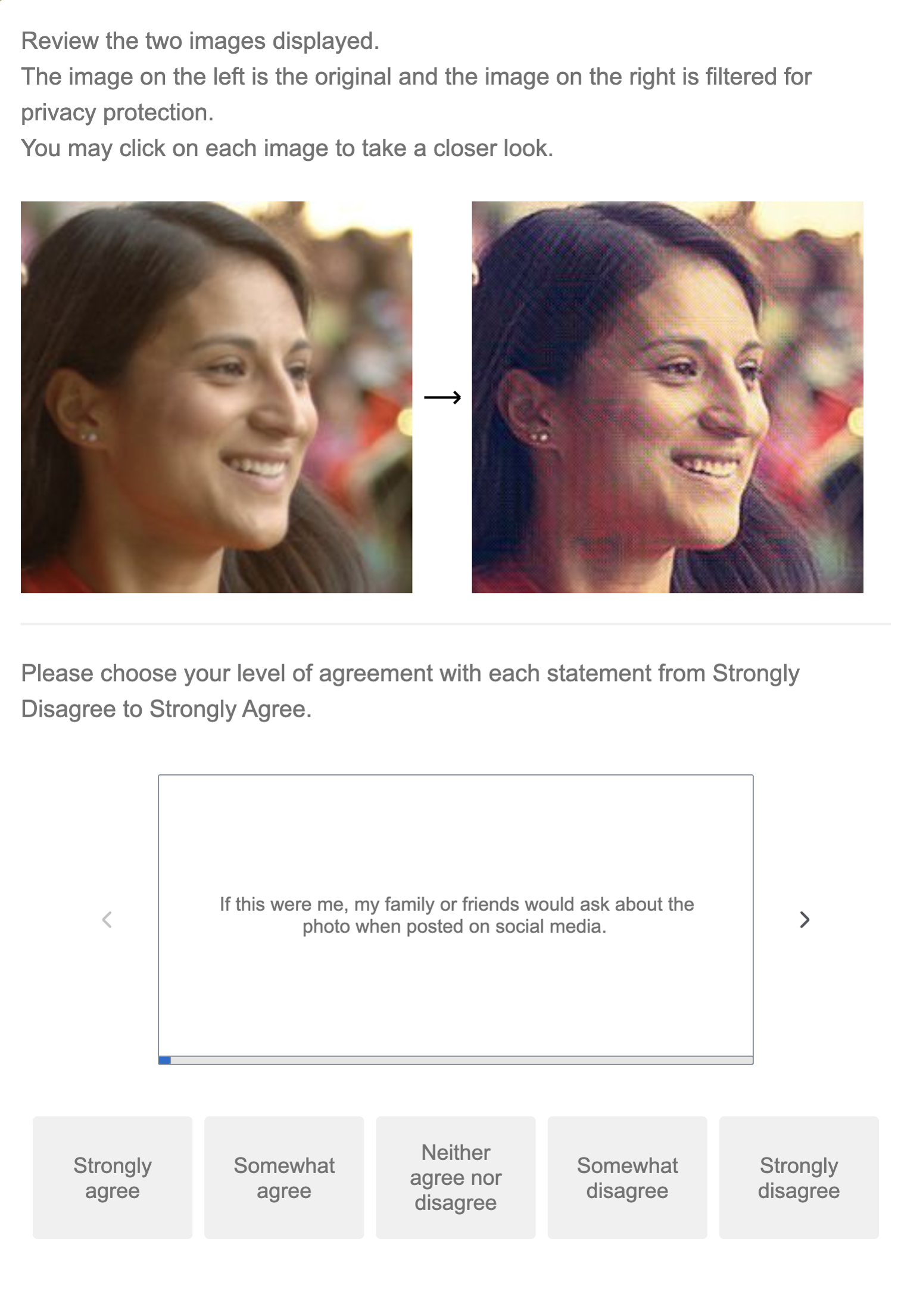}
    \caption{Screenshot of survey instrument used for collection of \acrshort{saia8} responses.}
    \label{fig:survey}
    \Description[A comparison task showing an original photograph next to a privacy-filtered version, followed by a survey question asking about social media sentiment.]{Image Comparison and Survey Task. This interface presents a visual comparison between two images side-by-side, separated by an arrow indicating transformation. The image on the left is the original photograph of a smiling woman. The image on the right is the same photo that has been filtered for privacy protection. Below the images is a survey prompt requiring user feedback: ``If this were me, my family or friends would ask about the photo when posted on social media.'' Users are required to choose their level of agreement using a standard five-point Likert scale, which includes options for Strongly agree, Somewhat agree, Neither agree nor disagree, Somewhat disagree, and Strongly disagree.}
\end{figure}

\subsection{Survey Procedure}
Participants were recruited from Prolific\footnote{\url{https://prolific.com}} and paid \$0.5 for a response, which took two minutes on average to complete.
We had each participant respond to the \acrshort{saia8} for an output generated by only one of the seven defenses, a between-subjects approach, to avoid order effects in the response.
After informing the participant about the study and collecting consent per our IRB-approved study design, the participant was shown a side-by-side of an undefended image and the protected variant (matching the defense to which they were assigned, see \ref{fig:survey}).
Then, we asked participants to consider the two images while responding to the \acrshort{saia8}, presented in randomized order.
% for a single image and a single obfuscation approach to mitigate the impact of the image content and ordering effect -- i.e., between subjects
% Second, participants are asked to select up to three of the fourteen obfuscation techniques -- Figure \ref{fig:obfuscation-examples} -- that they would consider using.

% In the next section of our study, we presented participants with the same image protected by our eight aesthetic filters, Fawkes, LowKey. We asked users to select up to three of the thirteen images that they would share to social media. Users were able to select no images if they did not find any filters acceptable.

\subsection{Study Iterations}
\begin{figure*}
    \centering
    \begin{tabular}{ccccccc}
        \toprule
        Control & Fawkes & LowKey & Dogpatch & Moon & Nashville & Sutro \\ \midrule
        \includegraphics[width=1.75cm]{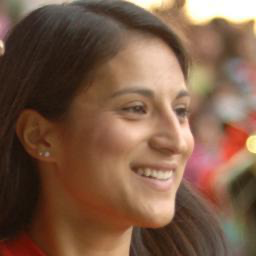} &
            \includegraphics[width=1.75cm]{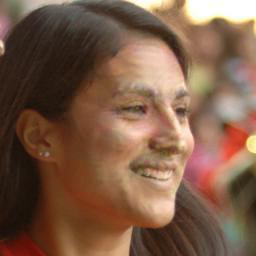} &
            \includegraphics[width=1.75cm]{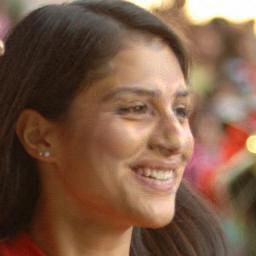} &
            \includegraphics[width=1.75cm]{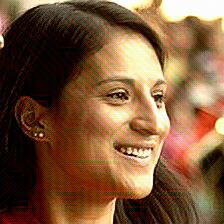} &
            \includegraphics[width=1.75cm]{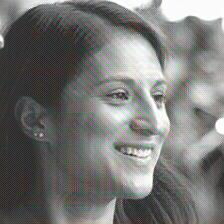} &
            \includegraphics[width=1.75cm]{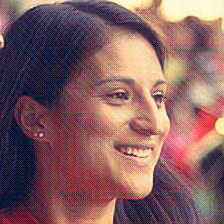} &
            \includegraphics[width=1.75cm]{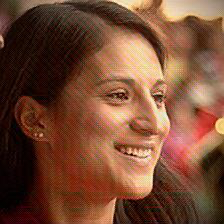} \\

        \includegraphics[width=1.75cm]{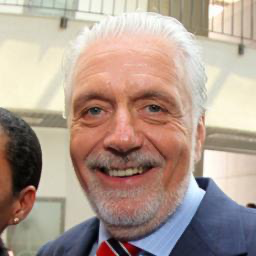} &
            \includegraphics[width=1.75cm]{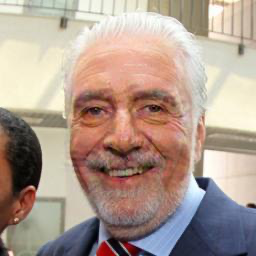} &
            \includegraphics[width=1.75cm]{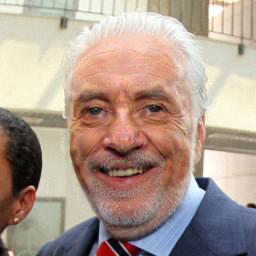} &
            \includegraphics[width=1.75cm]{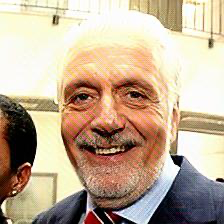} &
            \includegraphics[width=1.75cm]{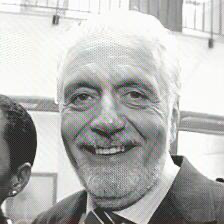} &
            \includegraphics[width=1.75cm]{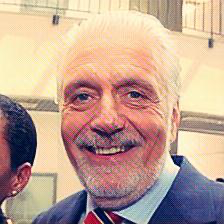} &
            \includegraphics[width=1.75cm]{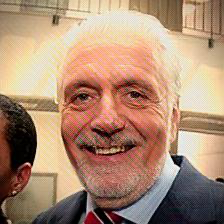} \\

        \includegraphics[width=1.75cm]{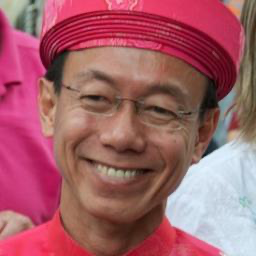} &
            \includegraphics[width=1.75cm]{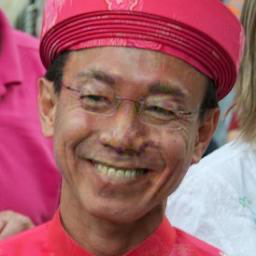} &
            \includegraphics[width=1.75cm]{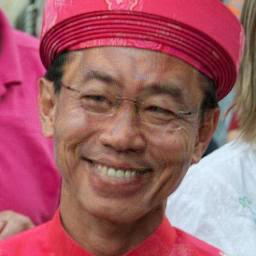} &
            \includegraphics[width=1.75cm]{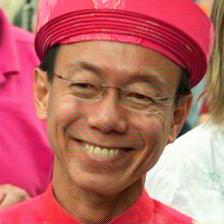} &
            \includegraphics[width=1.75cm]{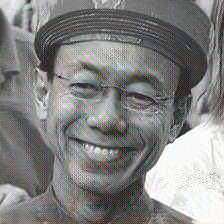} &
            \includegraphics[width=1.75cm]{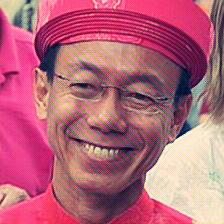} &
            \includegraphics[width=1.75cm]{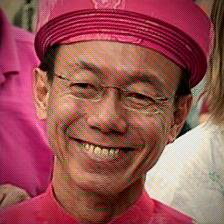} \\

        \includegraphics[width=1.75cm]{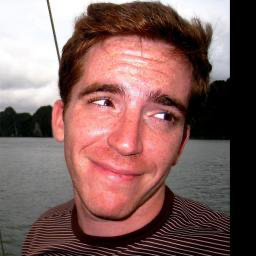} &
            \includegraphics[width=1.75cm]{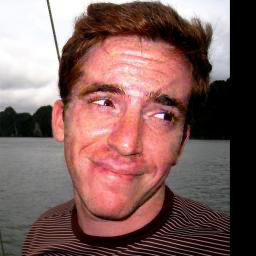} &
            \includegraphics[width=1.75cm]{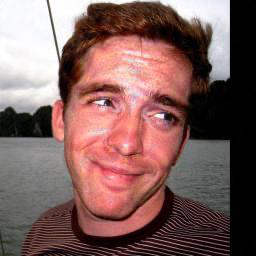} &
            \includegraphics[width=1.75cm]{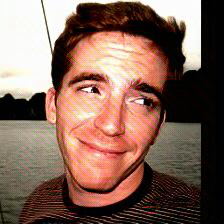} &
            \includegraphics[width=1.75cm]{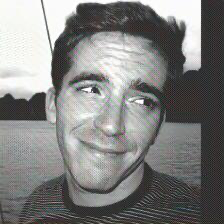} &
            \includegraphics[width=1.75cm]{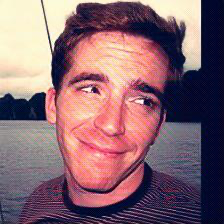} &
            \includegraphics[width=1.75cm]{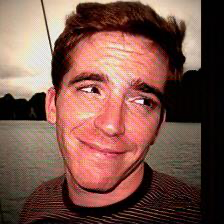} \\

        \includegraphics[width=1.75cm]{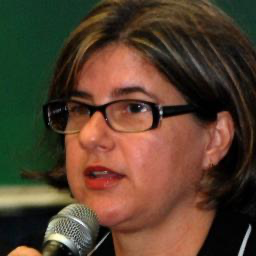} &
            \includegraphics[width=1.75cm]{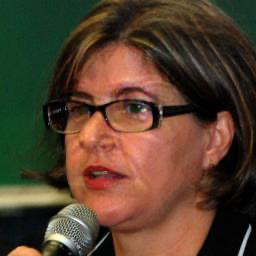} &
            \includegraphics[width=1.75cm]{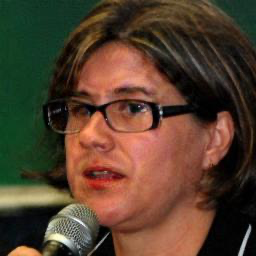} &
            \includegraphics[width=1.75cm]{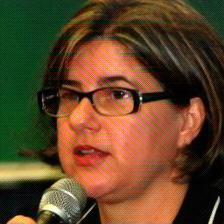} &
            \includegraphics[width=1.75cm]{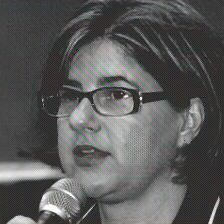} &
            \includegraphics[width=1.75cm]{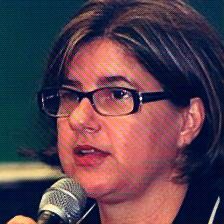} &
            \includegraphics[width=1.75cm]{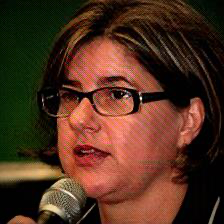} \\
    \end{tabular}
    \caption{Five images selected for second iteration of \acrshort{saia8} data collection.}
    \label{fig:wave2images}
\end{figure*}

We ran two separate iterations of this study, both with a sample of $N=315$ participants based on a power analysis.
% \footnote{See power analysis in Figure \ref{tab:powanalysis}}.
Each iteration differed only in base image selection criteria.
In the first iteration, we randomly selected a base image from the 1K \acrshort{fdf} subset described above.
In theory, random selection from this large set of images should mitigate any impact that image content or quality has on \acrshort{saia8} reports.
However, given that the \acrshort{fdf} dataset includes blurry, dark, occluded, and non-human faces, we conducted a second iteration of the study with tighter controls on image content to ensure that the presented images had a clearly visible face.
As a result, in the second iteration, we pre-selected five images from the 1K subset that contain forward facing, well lit, and human faces (see Figure \ref{fig:wave2images}).

% \begin{figure}[ht]
%     \centering
%     \subfloat[Fawkes]{\includegraphics[width=2.5cm]{figs/Wave2/Fawkes/example1.png}}
%     \hfil
%     \subfloat[LowKey]{\includegraphics[width=2.5cm]{figs/Wave2/LowKey/example2.png}}
%     \hfil
%     \subfloat[Sutro]{\includegraphics[width=2.5cm]{figs/Wave2/Sutro/example3.png}}
%     \hfil
%     \subfloat[Moon]{\includegraphics[width=2.5cm]{figs/Wave2/Moon/example4.png}}
%     \hfil
%     \subfloat[Nashville]{\includegraphics[width=2.5cm]{figs/Wave2/Nashville/example5.png}}
%     \hfil
%     \subfloat[Dogpatch]{\includegraphics[width=2.5cm]{figs/Wave2/Dogpatch/example2.png}}
%     \caption{Sampling of defenses applied to the five images selected for second iteration of \acrshort{saia8} data collection.}
%     \label{fig:wave2images}
% \end{figure}

\subsection{Analytical Approach}
We analyzed data from both iterations with the same analytical approach.
The \acrshort{saia8} data from each iteration of the first task was analyzed with a one-way \acrfull{anova} \cite{st_analysis_1989} which tests for differences in means of the dependent variable amongst groups in the independent variable with significance achieved when $\rho \leq 0.05$ or $\rho \leq 0.01$.
We followed this up with a post-hoc pair-wise Tukey t-test \cite{tukey_comparing_1949} to measure statistical differences amongst the pairs of all seven conditions.
This test is optimal for balanced groups -- i.e., the same number of responses per condition -- and provides a deeper understanding of which groups are statistically similar with respect to the dependent variable -- i.e., \acrshort{saia8}.
% Note that in our second iteration, we introduced a covariate -- the base image -- and thus include an analysis of how the base image may have impacted overall acceptability.

We fitted a Plackett-Luce \cite{turner_plackettluce_2025,turner_modelling_2020} model to assess defense output preference in the grid selection task and used quasi-variance calculations (using the R qvcalc package) to obtain standard errors for all pairwise comparisons.
The worth parameter represents each filter’s relative preference strength on a log scale, where higher values indicate greater preference.

\subsection{Results}
\subsubsection{Acceptability (SAIA-8) Results}
In our analysis of the data from both iterations of \acrshort{saia8} responses, we found statistical significance both in measurable differences amongst all the groups and between pairs -- see \ref{tab:pair-obfs}.
% In our analysis of the data from both iterations of \acrshort{saia8} responses, we found statistical significance both in measurable differences amongst all the groups -- see \ref{tab:anova} -- and between pairs -- see \ref{tab:pair-obfs}.
In the both iterations, the \acrshort{anova} test showed a significant difference in the \acrshort{saia8} scores amongst the seven conditions ($\mathbf{\rho_{1}=1.53 \times 10^{-18}}$ and $\mathbf{\rho_{2}=2.68 \times 10^{-16}}$).
We also measured the effect size for these conditions ($\eta_{1}^{2} = 0.27$ and $\eta_{2}^{2} = 0.24$).
Thus, our sample size of $N=315$ per iteration with seven groups and a targeted significance of $\alpha=0.01$ achieved an approximate power of $1.0$.

To better understand the statistical relationships between defenses, we performed a post-hoc Tukey t-test \cite{tukey_comparing_1949}.
We summarized the results in Table~\ref{tab:tukey-reduced-saia} for analysis between \textsc{AuraMask}-based defenses and prior approaches -- i.e., Fawkes \cite{shan_fawkes_2020} and LowKey \cite{cherepanova_lowkey_2021}.
% \footnote{For pairwise differences between AuraMask-based defenses see Appendix~\ref{tab:tukey-full-saia}}.
% \input{tabs/tukey-groups}
\begin{figure*}[h]
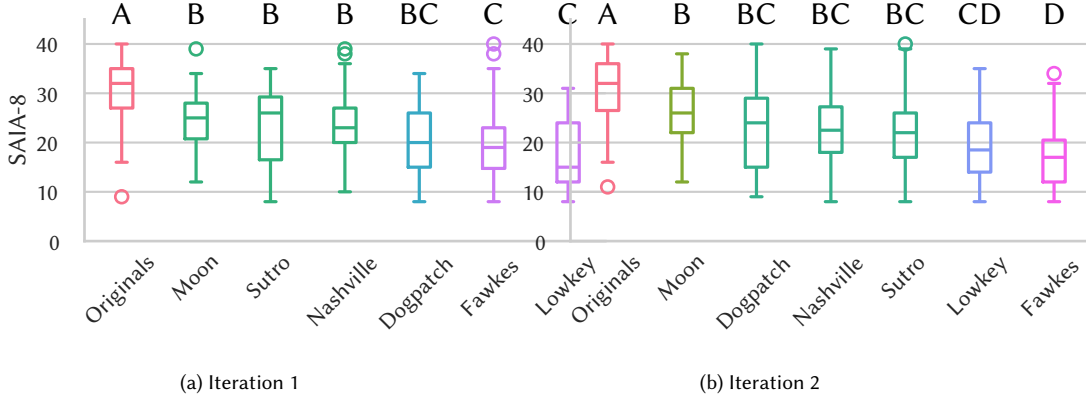

    \centering
    \begin{subfigure}{0.45\textwidth}
        \input{figs/TukeyBox/boxplot_w1.pgf}
        \caption{Iteration 1}
        \label{fig:iter1-tukeybox}
    \end{subfigure}
    \begin{subfigure}{0.45\textwidth}
        \input{figs/TukeyBox/boxplot_w2.pgf}
        \caption{Iteration 2}
        \label{fig:iter2-tukeybox}
    \end{subfigure}
    \caption{Distribution of \acrshort{saia8} scores in each iteration with conditions grouped on statistical similarity ($\rho > 0.05$).}
    \label{fig:box_tukey}
\end{figure*}
Of note in these results is the \textsc{AuraMask}-based defense that emulates the Moon filter.
This black-and-white filter achieved significantly higher scores than both baselines across the two study iterations.
The other \textsc{AuraMask}-based defenses we tested also achieved higher \acrshort{saia8} scores than the baseline (see Figure \ref{fig:box_tukey}), but the statistical significance of these differences varied across iteration.
In iteration 1, Moon, Nashville and Sutro scored significantly higher than both Fawkes and LowKey, but Dogpatch did not.
In iteration 2, all of our defenses scored significantly higher than Fawkes, but only Moon scored significantly higher than LowKey.
Here, we use \acrfull{cld} to identify groups of pairs based on significance level -- i.e., grouping for pairwise significance $\rho > 0.05$ (see \ref{fig:box_tukey}).
We also present raw statistical data of the pairwise test between the control group and obfuscation groups (see Table \ref{tab:tukey-reduced-saia}).
% \footnote{Full pairwise results available in Table \ref{tab:tukey-w1} and Table \ref{tab:tukey-w2}}.

\begin{table}[ht]
    \centering
    \begin{tabular}{cccccc}
        \toprule
                                                                                            &                    & \multicolumn{2}{c}{Iteration 1}                                      & \multicolumn{2}{c}{Iteration 2}                                \\ 
        A                                                                                   & B                  & $\rho$                        &  $\Delta$                            & $\rho$                         & $\Delta$                      \\ \cmidrule(lr){1-2} \cmidrule(lr){3-4} \cmidrule(lr){5-6}
        \multirow{4}{*}{\parbox[t]{2mm}{\rotatebox[origin=c]{90}{\textit{Fawkes}}}}         & \textit{Dogpatch}  & $0.88$                        & $-1.72$                              & $\mathbf{4.65\mathrm{e}{-3}}$  & $-5.55$                       \\
                                                                                            & \textit{Moon}      & $\mathbf{3.18\mathrm{e}{-4}}$ & $-5.73$                              & $\mathbf{4.49\mathrm{e}{-7}}$  & $-8.68$                       \\
                                                                                            & \textit{Nashville} & $1.98\mathrm{e}{-2}$          & $-4.42$                              & $\mathbf{9.75\mathrm{e}{-3}}$  & $-5.32$                       \\
                                                                                            & \textit{Sutro}     & $1.99\mathrm{e}{-2}$          & $-4.59$                              & $3.66\mathrm{e}{-2}$           & $-4.66$                       \\ \cmidrule(lr){3-4} \cmidrule(lr){5-6}
        \multirow{4}{*}{\parbox[t]{2mm}{\rotatebox[origin=c]{90}{\textit{LowKey}}}}         & \textit{Dogpatch}  & $0.12$                        & $-3.88$                              & $0.19$                         & $-6.72$                       \\
                                                                                            & \textit{Moon}      & $\mathbf{6.99\mathrm{e}{-7}}$ & $-7.89$                              & $\mathbf{2.04\mathrm{e}{-4}}$  & $-3.36$                       \\
                                                                                            & \textit{Nashville} & $\mathbf{1.28\mathrm{e}{-4}}$ & $-6.58$                              & $0.28$                         & $-2.70$                       \\
                                                                                            & \textit{Sutro}     & $\mathbf{1.52\mathrm{e}{-4}}$ & $-6.75$                              & $0.55$                         & $-3.13$                       \\ \bottomrule
    \end{tabular}
    \caption{Pairwise analysis between \acrshort{saia8} scores for defenses A and B with ($\rho \leq 0.01$) results bolded.} %Full results in Table \ref{tab:tukey-full-saia}.%}
    \label{tab:tukey-reduced-saia}
\end{table}

\begin{figure*}[h]
    \centering
    \input{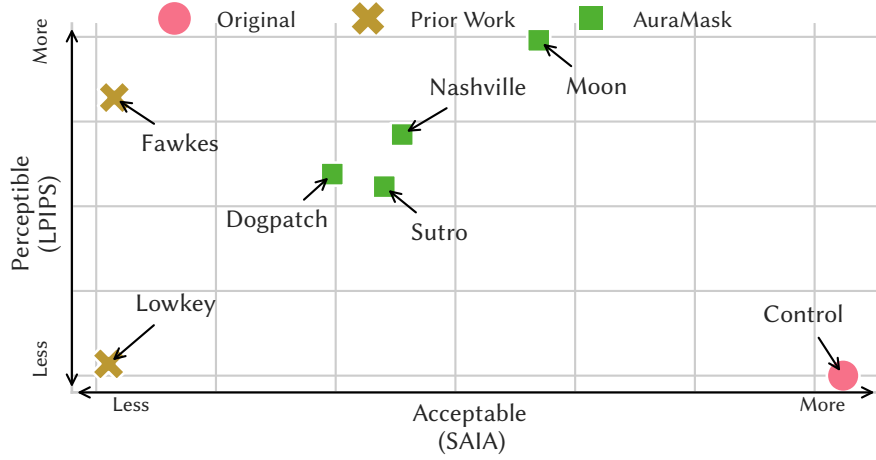}
    \caption{Defenses tested in the acceptability task, charted along user acceptability (x-axis) and similarity relative to the original image (y-axis). The Moon filter, despite being the most perceptually distinctive from the original (Control) image, was also rated as most acceptable amongst the obfuscated images users rated.}
    \label{fig:the_graph}
\end{figure*}

\textit{Perceptual similarity vs. Acceptability:} In Figure~\ref{fig:the_graph}, we plot overall the overall user acceptability of defenses against their perceptual similarity. LowKey has the highest perceptual similarity with the original unaltered image, but is amongst the least acceptable of the options we tested. In contrast, the Moon defense is both the most perceptually distinct from the original image and also the most user acceptable. In short, aiming for ``perceptibly'' aesthetic defenses seems more prudent for user acceptance than aiming for ``imperceptibility'' as has been the focus of prior approaches. 

\subsubsection{Preference (Grid Selection) Results}
\begin{figure*}[h]
    \centering
    \includegraphics[width=0.8\textwidth]{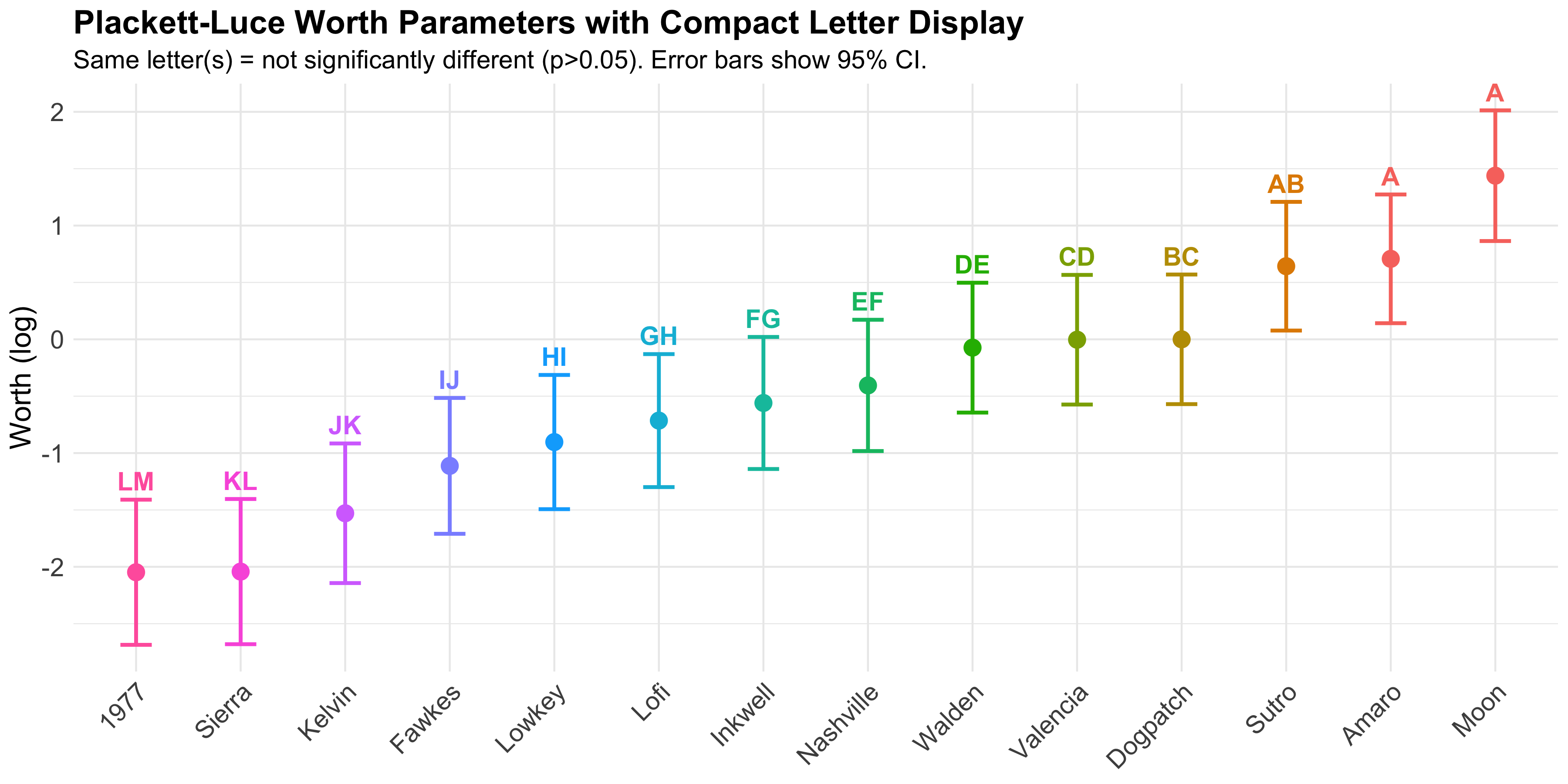}
    \caption{Plackett-Luce worth parameters for the grid preference selection task, grouped using pairwise statistical comparisons. Higher values indicate greater preference. Options that do not share a letter are significantly different; those that do share a letter are not.}
    \label{fig:acceptance}
\end{figure*}

Our analysis revealed distinct preference tiers within the grid-selection task responses.
To identify groups of statistically equivalent defenses, we constructed a \acrshort{cld} using pairwise comparisons at $\alpha = 0.05$.
In the \acrshort{cld}, only filters with no shared letters are significantly different.
For example, filters labeled ``AB'' in Figure~\ref{fig:acceptance} form a bridge between groups A and B, being statistically indistinguishable from both groups whereas ``A'' and ``B'' are statistically distinguishable from each other.
The Inkwell through Moon \textsc{AuraMask} defenses (\acrshort{cld} groups A to FG) were all significantly more preferred than Fawkes and Lowkey (groups IJ and HI, respectively).
At the lower end of the preference spectrum, 1977 and Sierra (groups LM and KL) were significantly less preferred than the majority of filters.
The top-tier defenses -- Moon, Amaro, and Sutro (groups A and AB) -- showed the strongest preferences, with Moon having the highest worth parameter (log-worth = $1.44$, worth = $4.21$).

% We next present the rate at which users selected different filters in Figure~\ref{fig:acceptance} in the second task (grid selection). We found that users were more likely to select eight of our eleven filters, which use visible but aesthetic modifications, over LowKey and Fawkes. 
% Of note, the Moon filter --- which also scored highest on the \acrshort{saia8} scale --- was selected by 12\% of our participants, more than twice the rate of Fawkes (4\%) and LowKey (5\%).

In all, our results demonstrate that \textsc{AuraMask}-based defenses were significantly more user acceptable and preferable than Fawkes and LowKey.

% Overall, this study provides empirical evidence regarding the relative acceptability between the AuraMask defense approach and prior ``imperceptible'' defenses.

%These results show that giving users a choice of how their photos are modified to protect against facial recognition, including providing them with a range of visible but aesthetic AFR modifications, le
%Additionally, we find that providing users a choice of how their photos are modified  to protect against facial recognition leads to more users choosing any AFR filter.

% \begin{figure*}[h]
%     \centering
%     \input{figs/selection.pgf}
%     \caption{Counts at which users choose different AFR filters when asked to select up to three filters they would post to social media with.}
%     \label{fig:acceptance}
% \end{figure*}

% \section{Results}
% \input{content/Results.tex}

\section{Discussion}
In this work, we sought to uncover \emph{how we can design anti-facial recognition defenses to better align with people's aethetic and self-presentation preferences}.
We introduced an extensible pipeline --- \textsc{AuraMask} --- which aimed to ease development of \acrshort{afr} defenses that are both adversarially effective and improve user acceptance.
Through \textsc{AuraMask}, we generated a set of 80 novel defenses -- 40 \acrfull{singletarget} and 40 \acrfull{ensembletarget} -- that serve as a proof-of-concept (\emph{RQ1}) for incorporating ``aesthetics'' into \acrshort{afr} defenses.
% Using \textsc{AuraMask}, we generated a set of such filters that successfully minimize face verification effectiveness and emulate the 40 aesthetic image filters available on Instagram.
In our comparative technical evaluation (\emph{RQ2}), we demonstrated strong technical performance. The \acrshort{ensembletarget}s outperformed previous methods -- Fawkes \cite{shan_fawkes_2020} and LowKey \cite{cherepanova_lowkey_2021} -- on all tested facial recognition models, while the \acrshort{singletarget}s did the same against the targeted model (but were weaker against untargeted models).
Moreover, our user-study (\emph{RQ3}) suggests that users find ``aesthetic'' defenses more acceptable than and preferable to prior work.
% In the design and evaluation process, we arrived at multiple insights on the ``strength'' of these obfuscations, potential pitfalls in their use, and the path forward for anti-facial recognition obfuscations.

\subsection{Are AFR defenses ethical and appropriate?}\label{sec:discussion_protection}
There is debate within the \acrlong{sandp} community on whether user-facing \acrshort{afr} technologies is appropriate to address facial recognition harms. Critics have argued that \acrshort{afr} defenses fail to provide strong protective guarantees and may thus endanger users by creating a false sense of security \cite{radiya-dixit_data_2021}.

This critique is valid, but assumes a traditional security model where defenders possess power and resources equal to or greater than assumed attackers. We argue that \acrshort{afr} filters should instead be understood as \textit{resistance technologies} \cite{agnew_technologies_2023}: tools deployed by under-powered groups (e.g., individual citizens) to protect themselves from powerful threat actors (e.g., governments and corporations using facial recognition). When defenders are under-resourced relative to attackers, absolute guarantees of protection are often neither possible nor reasonable. Instead, following Brunton and Nissenbaum's framework of obfuscation as a ``weapon of the weak'' \cite{brunton_obfuscation_2015}, we evaluate these technologies against different goals: providing cover, buying time, enabling deniability, and expressing protest.

Consider a user, Johnny, who wishes to maintain both a professional online presence and participate in political activism. Even if institutions or third-parties capture unprotected reference images of Johnny through CCTV or bystander photography, aesthetic \acrshort{afr} filters still offer meaningful protection across multiple dimensions:

\textbf{Providing Cover}: When Johnny applies obfuscation defenses to his online images, unobfuscated reference images captured in the physical world will have low probability of matching against his obfuscated posts. This mismatch provides cover for his online activities, particularly on platforms where he engages in activism under pseudonyms.

\textbf{Buying Time}: While sophisticated adversaries may possess countermeasures (or created obfuscated reference images to break these obfuscations), doing so requires individual analyst attention and scrutiny. This added friction reduces the effectiveness of mass surveillance systems designed to process millions of faces automatically, buying Johnny and others time before their accounts are identified and monitored.

\textbf{Deniability}: Unlike prior defenses that introduce artifacts unaligned with human aesthetic preferences (e.g., Fawkes \cite{shan_fawkes_2020}) or destructive obfuscations (e.g., blurring), aesthetic defenses provide plausible deniability. If confronted about images obfuscated with \textsc{AuraMask} defenses, Johnny can credibly claim he simply likes how the filters look. This aesthetic plausibility distinguishes our approach from methods whose protective intent is undeniable.

\textbf{Expressing Protest}: Beyond individual protection, widespread adoption of \acrshort{afr} filters serves as measurable resistance against surveillance systems. As ACLU lawyer Ashley Gorski argues, quantifying how many people take evasive action against surveillance provides concrete evidence of harm when advocating for legal redress \cite{gorski_biden_2022}. Each visibly filtered image becomes both protection and protest.

These goals represent meaningful victories for under-powered defenders, even if they do not provide absolute guarantees.
% As with historical examples of obfuscation in resistance movements (e.g., Operation Vula in South Africa), the value lies not in perfect security but in being ``strong enough'' for the context and goals at hand \cite{brunton_obfuscation_2015}.
Thus, as long as users receive \textit{adequate risk communication} about the limitations and strengths of these tools, and use these tools as one of a broader suite of protection strategies \cite{das_subversive_2020}, we argue that developing and deploying resistance technologies such as aesthetic \textsc{AuraMask} defenses remains both necessary and important.

\subsection{Can two AFR-filtered images be correctly matched to one another?}
% The face verification task operates by drawing inferences based on clustering of data points within some $N$-dimensional space -- i.e., \emph{face embeddings}.
% While the AuraMask losses we propose -- \acrshort{fea} and \acrshort{feat} -- are effective in shifting embeddings outside of the original cluster, it is possible these optimizations simply moves the cluster location in $N$-dimensional space.

\textsc{AuraMask} makes it so an obfuscated photo of Johnny isn't recognized as being the same person as an unobfuscated photo of Johnny.
But what about two obfuscated photos of Johnny? Would they both be identifiable as the same Johnny?
In the context of an institution searching for social media presence, Johnny may be identifiable if the reference image is obfuscated with the same protective filter.
Indeed, Figure~\ref{fig:both-obfuscated} demonstrates this concern as a distinct possibility as face verification recall improves when comparing two images protected with the same defense.
While an adversary can themselves apply an obfuscation to a reference image, consider how in this paper we present 40 unique defenses  -- 80 if we consider both configurations -- with different visual attributes and that the \textsc{AuraMask} pipeline allows for the creation of boundless numbers of other such defenses.
This provides several choices for obfuscation before sharing, all of which are trained separately and are unlikely to move to the same position in the $N$-dimensional face embedding space.
Figure~\ref{fig:mixed-obfuscated} illustrates how choice in obfuscation could provide greater protection against facial recognition systems even when both images are protected.
The large number of defenses that can be created with \textsc{AuraMask} points the way to future ``one-time pad'' style obfuscations that users can dynamically apply.
We note, however, that the security of this scheme should be tested.

\subsection{New Horizons for ``Aesthetically Perceptible'' Obfuscation}
We began this work by detailing the \textsc{AuraMask} pipeline for training multi-task \acrshort{afr} defenses and describing how we used it to generate Instagram filter-like defenses.
These defenses were utilized to investigate the impact of relaxing the perceptibility constraint in both user acceptance and technical efficacy.
In our evaluations, we found our defenses were successful in sabotaging face verification and achieved higher measures of user acceptance than approaches that optimized imperceptibility.
However, there remains much to explore in this design space and the defenses we presented here are only the ``tip of the iceberg.''
Indeed, the evaluated obfuscations have room for improvement -- e.g., \acrshort{ensembletarget}s are more protective but introduce structurally severe perturbations (see Figure~\ref{fig:obfuscation-examples}).
% -- tunable through nearly infinite permutations of hyperparameters -- e.g., embedding model targets, \acrshort{atn} architectures, training datasets, pre- and post-processing steps, and so on.
Moreover, the ``aesthetic'' heuristics we use in this work are simplistic when compared to other methods in computer vision and computational photography -- e.g., style transfer \cite{bui_structureaware_2022,yim_filter_2020}, heuristics for image aesthetics \cite{talebi_nima_2018,chen_topiq_2023}, or emulating expert adjustments \cite{ho_deep_2021,bychkovsky_learning_2011,tseng_neural_2022} -- which can go beyond simple emulation.
More artistically-inclined developers may find ways to incorporate \acrshort{aml} perturbations in ways we have not considered.
In short, eschewing the perceptibility constraint from \acrshort{aml} research unlocks the potential for a large ecosystem of creative \acrshort{afr} defenses that may be accelerated by the \textsc{AuraMask} toolkit.
% In short, adopting the imperceptibility constraint from \acrshort{aml} research severely limited the potential for creative and accepted obfuscations; in this work, we demonstrate that even a minor shift toward intentionally perceptible obfuscations improves acceptance without sacrificing protection.
% It is time for those developing user-facing \acrshort{aml} obfuscations to eschew imperceptibility.

\section{Limitations}
First, in our user study, we only used the \acrshort{singletarget}s that have less obfuscation strength than \acrshort{ensembletarget}.
We opted for \acrshort{singletarget}s as \acrshort{ensembletarget} outputs introduced more structural alterations that alter the nose or introduce ``ghost'' features around the subject (see Figure \ref{fig:obfuscation-examples}).
While the \acrshort{ensembletarget} configuration outperforms prior systems and successfully emulates Instagram filters, we assumed the structural artifacts would not align with preferred presentation.
We leave tuning of the Ensemble configuration for less structural alterations to future work.

Second, participants were asked to respond to the \acrshort{saia8} with pre-obfuscated images of third-party subjects as opposed to personal photos, potentially biasing scores.
We took this approach as the \acrshort{pgd}-based approaches -- \cite{shan_fawkes_2020} and \cite{cherepanova_lowkey_2021} -- can take up to 15 minutes, likely dis-incentivizing participation.
Furthermore, we expect that requiring participants to upload a personal photo would potentially bias our sample in other ways --- many privacy-conscious participants may be apprehensive.
Further evaluation is warranted, especially in a field context with users obfuscating personal photos.
Moreover, given our findings here, perhaps these \acrshort{pgd} methods no longer need to be considered in future evaluations.

Finally, neither Fawkes nor LowKey remain state-of-the-art with several ostensibly user-facing obfuscations published since \cite{hussain_reface_2023, chandrasekaran_faceoff_2021,wenger_sok_2023}.
Unfortunately, many of these newer obfuscations have not open-sourced their code; Fawkes and LowKey, to their great credit, have more accessible codebases.
For our part, we freely provide not only the \textsc{AuraMask} source-code but also pre-trained models for comparative evaluation at \url{https://hf.co/collections/logasja/auramask}.

\section{Conclusion}
In this work, we present a novel toolkit -- \textsc{AuraMask} -- that can create \acrshort{afr} image defenses that are both adversarially effective and aesthetically pleasing. We used \textsc{AuraMask} to generate 80 ``aesthetic'' \acrshort{afr} defenses for a technical and user-centered evaluation.
% These proof-of-concept defenses captured ``aesthetics'' by emulating well-known Instagram image filters.
Our experimental results demonstrated that \textsc{AuraMask} defenses achieve similar -- and at times better -- protection against face verification than current baselines, reducing face verification recall by as much as $99\%$.
In addition, our user-study results demonstrated that the output of \textsc{AurMask} defenses achieve significantly higher user acceptance over prior work, and were more frequently preferred than baseline methods.
One defense of note emulated the ``Moon'' Instagram filter, achieving the highest user preference and performing best against all facial recognition models tested.
\textsc{AuraMask} enables the exploration of infinitely many new ``aesthetic'' \acrshort{afr} defenses that may be more in line with users' self-presentation preferences and thus more likely to see widespread use. To help accelerate this exploration by those both within and outside \acrfull{sandp}, we fully open source \textsc{AuraMask}.
% In all, we show that the potential benefits in user acceptance and increased design space of intentionally perceptible \acrshort{afr} defenses deserves further exploration by those within and outside \acrfull{sandp} which we hope the \textsc{AuraMask} pipeline
\footnote{Available here: \url{https://gitlab.com/raccs-lab/auramask-library}}

\begin{acks}
This work was supported, in part, by the National Science Foundation (NSF) under
SaTC Award No. 2316287.
\end{acks}

\bibliographystyle{ACM-Reference-Format}
\bibliography{references}

\appendix
% \printglossary[type=\acronymtype]

\begin{table}[ht]
    \centering
    \begin{tabular}{p{0.01\columnwidth}p{0.85\columnwidth}}
        \toprule
        1 & I don't feel comfortable with the changes made to the photograph.                       \\ 
        2 & I feel concerned with how the filter has affected my looks.                             \\ 
        3 & I feel the filtered photo's changes are immediately noticeable.                         \\ 
        4 & My family or friends would ask about the filtered photo if I posted it on social media. \\ 
        5 & The filter makes me look less human.                                                    \\ 
        6 & The changes made by the filter defeat the purpose of sharing the image.                 \\ 
        7 & I wouldn't share the image publicly after the filter was applied.                       \\ 
        8 & I would rarely use this filter for photos shared to social media.                       \\ \bottomrule
    \end{tabular}%
    \caption{The \acrshort{saia8} questionnaire items. Responses are on a 5-point Likert scale from Strongly Disagree (1) to Strongly Agree (5).}
    \label{appx:saia-8}
\end{table}

% \begin{table}[ht]
%     \centering
%     \begin{tabular}{cccc}
%         \toprule
%         Epoch   & Output                                               & TopIQ  & \acrshort{feat}      \\ \midrule
%         100     & \includegraphics[width=9em]{figs/TopIQOnly/100.png}  & $0.151$  & $-0.224$           \\
%         % 200     & \includegraphics[width=9em]{figs/TopIQOnly/200.png}  & $0.129$  & $-0.237$           \\
%         300     & \includegraphics[width=9em]{figs/TopIQOnly/300.png}  & $0.123$  & $-0.246$           \\
%         % 400     & \includegraphics[width=9em]{figs/TopIQOnly/400.png}  & $0.117$  & $-0.259$           \\
%         500     & \includegraphics[width=9em]{figs/TopIQOnly/500.png}  & $0.116$  & $-0.258$           \\ \bottomrule
%     \end{tabular}
%     \caption{\acrshort{atn} trained only using TopIQ as perceptual loss.}
%     \label{fig:topiq-only}
% \end{table}

\begin{figure}[ht]
    \centering
    \begin{tabular}{cccccc}
        \toprule
        \multicolumn{6}{c}{Output at Epoch} \\ \cmidrule(lr){3-4}
        \multicolumn{2}{c}{100}                             & \multicolumn{2}{c}{300}                               & \multicolumn{2}{c}{500} \\
        \multicolumn{2}{c}{\includegraphics[width=6em]{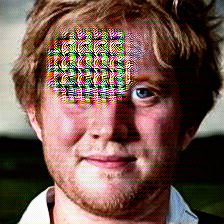}} & \multicolumn{2}{c}{\includegraphics[width=6em]{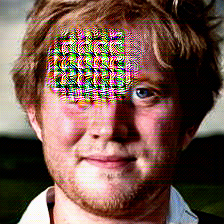}}   & \multicolumn{2}{c}{\includegraphics[width=6em]{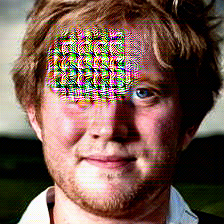}} \\ 
        TopIQ   & \acrshort{feat} & TopIQ   & \acrshort{feat} & TopIQ   & \acrshort{feat} \\ \cmidrule(lr){1-2} \cmidrule(lr){3-4} \cmidrule(lr){5-6}
        $0.151$ & $-0.224$ & $0.123$ & $-0.246$ & $0.123$ & $-0.258$ \\ \bottomrule
    \end{tabular}
    \caption{\acrshort{atn} trained only using TopIQ as perceptual loss.}
    \label{fig:topiq-only}
\end{figure}

\begin{table}[ht]
    \centering
    \begin{tabular}{@{}rrcccccc@{}}
        \toprule
        \multicolumn{1}{l}{}                    & \multicolumn{2}{c}{ArcFace}                                            & \multicolumn{2}{c}{Facenet}                                             & \multicolumn{2}{c}{VGGFace}                                               \\ 
        \multicolumn{1}{l}{}                    & \textit{\acrshort{singletarget}}  & \textit{\acrshort{ensembletarget}} &  \textit{\acrshort{singletarget}}  & \textit{\acrshort{ensembletarget}} & \textit{\acrshort{singletarget}}  & \textit{\acrshort{ensembletarget}}    \\ \cmidrule(l){2-3} \cmidrule(l){4-5} \cmidrule(l){6-7}
        Dogpatch-Dogpatch                       & $0.93$                            & $0.94$                             & $0.83$                             & $0.67$                             & $0.77$                            & $0.85$                                \\ \cmidrule(l){2-3} \cmidrule(l){4-5} \cmidrule(l){6-7}
        Nashville-Nashville                     & $0.85$                            & $0.94$                             & $0.82$                             & $0.58$                             & $0.81$                            & $0.80$                                \\ \cmidrule(l){2-3} \cmidrule(l){4-5} \cmidrule(l){6-7}
        Sutro-Sutro                             & $0.87$                            & $0.97$                             & $0.83$                             & $0.77$                             & $0.81$                            & $0.72$                                \\ \cmidrule(l){2-3} \cmidrule(l){4-5} \cmidrule(l){6-7}
        Dogpatch-Moon                           & $0.88$                            & $0.90$                             & $0.75$                             & $0.50$                             & $0.77$                            & $0.84$                                \\ \cmidrule(l){2-3} \cmidrule(l){4-5} \cmidrule(l){6-7}
        Dogpatch-Nashville                      & $0.87$                            & $0.93$                             & $0.83$                             & $0.57$                             & $0.77$                            & $0.78$                                \\ \cmidrule(l){2-3} \cmidrule(l){4-5} \cmidrule(l){6-7}  
        Dogpatch-Sutro                          & $0.89$                            & $0.06$                             & $0.83$                             & $0.10$                             & $0.77$                            & $0.11$                                \\ \cmidrule(l){2-3} \cmidrule(l){4-5} \cmidrule(l){6-7} 
        Moon-Nashville                          & $0.84$                            & $0.88$                             & $0.75$                             & $0.44$                             & $0.77$                            & $0.79$                                \\ \cmidrule(l){2-3} \cmidrule(l){4-5} \cmidrule(l){6-7}
        Moon-Sutro                              & $0.85$                            & $0.02$                             & $0.73$                             & $0.07$                             & $0.77$                            & $0.09$                                \\ \cmidrule(l){2-3} \cmidrule(l){4-5} \cmidrule(l){6-7}
        Sutro-Nashville                         & $0.85$                            & $0.09$                             & $0.81$                             & $0.12$                             & $0.78$                            & $0.16$                                \\ \bottomrule
    \end{tabular}
    \caption{Paired Obfuscation recall performance with the face verification task using \acrshort{lfw} pairs dataset across \acrshort{singletarget} and \acrshort{ensembletarget}.}
    \label{tab:pair-obfs}
\end{table}

% \input{tabs/tukey-saia-fawkes}
% \input{tabs/tukey-saia-lowkey}

% \input{tabs/tukey-w1.tex}
% \input{tabs/tukey-w2.tex}

% \input{tabs/fdf-full-accuracy}

% \input{tabs/lfw-full-accuracy}

% \usepackage{booktabs}
% \usepackage{multirow}
% \usepackage{longtable}
% Note: It may be necessary to compile the document several times to get a multi-page table to line up properly
\begin{longtable}[c]{@{}cccccccccccccc@{}}
\caption{Recall rate across defenses and evaluations.}
\label{tab:full-accuracy}\\
\toprule
                        &                      & \multicolumn{4}{c}{\textbf{ArcFace}}                                                                                                                                                                                            & \multicolumn{4}{c}{\textbf{VGGFace}}                                                                                                                                                              & \multicolumn{4}{c}{\textbf{Facenet}}                                                                                                                                                              \\* \midrule
\multicolumn{1}{l}{}    & \multicolumn{1}{l}{} & \multicolumn{2}{c}{FDF}                                                                         & \multicolumn{2}{c}{LFW}                                                                                                       & \multicolumn{2}{c}{FDF}                                                                         & \multicolumn{2}{c}{LFW}                                                                         & \multicolumn{2}{c}{FDF}                                                                         & \multicolumn{2}{c}{LFW}                                                                         \\* \cmidrule(l){3-6} \cmidrule(l){7-10} \cmidrule(l){11-14} 
\endfirsthead
\multicolumn{14}{c}%
{{\bfseries Table \thetable\ continued from previous page}} \\
\toprule
                        &                      & \multicolumn{4}{c}{\textbf{ArcFace}}                                                                                                                                                                                            & \multicolumn{4}{c}{\textbf{VGGFace}}                                                                                                                                                              & \multicolumn{4}{c}{\textbf{Facenet}}                                                                                                                                                              \\* \cmidrule(l){3-14}
\multicolumn{1}{l}{}    & \multicolumn{1}{l}{} & \multicolumn{2}{c}{FDF}                                                                         & \multicolumn{2}{c}{LFW}                                                                                                       & \multicolumn{2}{c}{FDF}                                                                         & \multicolumn{2}{c}{LFW}                                                                         & \multicolumn{2}{c}{FDF}                                                                         & \multicolumn{2}{c}{LFW}                                                                         \\* \cmidrule(l){3-6} \cmidrule(l){7-10} \cmidrule(l){11-14} 
                        &                      & \textit{\acrshort{singletarget}}              & \textit{\acrshort{ensembletarget}}              & \textit{\acrshort{singletarget}}                             & \textit{\acrshort{ensembletarget}}                             & \textit{\acrshort{singletarget}}              & \textit{\acrshort{ensembletarget}}              & \textit{\acrshort{singletarget}}              & \textit{\acrshort{ensembletarget}}              & \textit{\acrshort{singletarget}}              & \textit{\acrshort{ensembletarget}}              & \textit{\acrshort{singletarget}}              & \textit{\acrshort{ensembletarget}}              \\* \cmidrule(l){3-4} \cmidrule(l){5-6} \cmidrule(l){7-8} \cmidrule(l){9-10} \cmidrule(l){11-12} \cmidrule(l){13-14} 
\endhead
Baseline                &                      & \multicolumn{2}{c}{$1.000$}                                                                     & \multicolumn{2}{c}{$0.905$}                                                                                                   & \multicolumn{2}{c}{$1.000$}                                                                     & \multicolumn{2}{c}{$0.798$}                                                                     & \multicolumn{2}{c}{$1.000$}                                                                     & \multicolumn{2}{c}{$0.873$}                                                                     \\* \cmidrule(l){3-6} \cmidrule(l){7-10} \cmidrule(l){11-14}  
\multicolumn{2}{c}{LowKey}                     & \multicolumn{2}{c}{$0.974$}                                                                     & \multicolumn{2}{c}{$0.122$}                                                                                                   & \multicolumn{2}{c}{$0.970$}                                                                     & \multicolumn{2}{c}{$0.263$}                                                                     & \multicolumn{2}{c}{$0.470$}                                                                     & \multicolumn{2}{c}{$0.026$}                                                                     \\* \cmidrule(l){3-6} \cmidrule(l){7-10} \cmidrule(l){11-14} 
\multirow{3}{*}{Fawkes} & \textit{L}           & \multicolumn{2}{c}{$0.993$}                                                                     & \multicolumn{2}{c}{$0.859$}                                                                                                   & \multicolumn{2}{c}{$1.000$}                                                                     & \multicolumn{2}{c}{$0.721$}                                                                     & \multicolumn{2}{c}{$0.888$}                                                                     & \multicolumn{2}{c}{$0.594$}                                                                     \\
                        & \textit{M}           & \multicolumn{2}{c}{$0.921$}                                                                     & \multicolumn{2}{c}{$0.671$}                                                                                                   & \multicolumn{2}{c}{$0.998$}                                                                     & \multicolumn{2}{c}{$0.539$}                                                                     & \multicolumn{2}{c}{$0.643$}                                                                     & \multicolumn{2}{c}{$0.180$}                                                                     \\
                        & \textit{H}           & \multicolumn{2}{c}{$0.864$}                                                                     & \multicolumn{2}{c}{$0.565$}                                                                                                   & \multicolumn{2}{c}{$0.997$}                                                                     & \multicolumn{2}{c}{$0.501$}                                                                     & \multicolumn{2}{c}{$0.519$}                                                                     & \multicolumn{2}{c}{$0.090$}                                                                     \\* \cmidrule[5\cmidrulewidth](l){3-6} \cmidrule[5\cmidrulewidth](l){7-10} \cmidrule[5\cmidrulewidth](l){11-14}
                        &                      & \textit{\acrshort{singletarget}}              & \textit{\acrshort{ensembletarget}}              & \textit{\acrshort{singletarget}}                             & \textit{\acrshort{ensembletarget}}                             & \textit{\acrshort{singletarget}}              & \textit{\acrshort{ensembletarget}}              & \textit{\acrshort{singletarget}}              & \textit{\acrshort{ensembletarget}}              & \textit{\acrshort{singletarget}}              & \textit{\acrshort{ensembletarget}}              & \textit{\acrshort{singletarget}}              & \textit{\acrshort{ensembletarget}}              \\* \cmidrule(l){3-4} \cmidrule(l){5-6} \cmidrule(l){7-8} \cmidrule(l){9-10} \cmidrule(l){11-12} \cmidrule(l){13-14} 
\multicolumn{2}{c}{1977}                       & $0.007$                                       & $0.025$                                         & $0.011$                                                      & $0.086$                                                        & $1.000$                                       & $0.075$                                         & $0.773$                                       & $0.032$                                         & $0.960$                                       & $0.449$                                         & $0.768$                                       & $0.372$                                         \\
\multicolumn{2}{c}{Aden}                       & $0.007$                                       & $0.026$                                         & $0.014$                                                      & $0.078$                                                        & $1.000$                                       & $0.092$                                         & $0.813$                                       & $0.024$                                         & $0.996$                                       & $0.127$                                         & $0.834$                                       & $0.103$                                         \\
\multicolumn{2}{c}{Amaro}                      & $0.020$                                       & $0.049$                                         & $0.025$                                                      & $0.136$                                                        & $1.000$                                       & $0.110$                                         & $0.793$                                       & $0.030$                                         & $0.992$                                       & $0.357$                                         & $0.829$                                       & $0.251$                                         \\
\multicolumn{2}{c}{Brannan}                    & $0.004$                                       & $0.017$                                         & $0.020$                                                      & $0.167$                                                        & $1.000$                                       & $0.119$                                         & $0.790$                                       & $0.042$                                         & $0.962$                                       & $0.391$                                         & $0.773$                                       & $0.207$                                         \\
\multicolumn{2}{c}{Ginza}                      & $0.055$                                       & $0.034$                                         & $0.015$                                                      & $0.115$                                                        & $1.000$                                       & $0.171$                                         & $0.797$                                       & $0.040$                                         & $0.975$                                       & $0.208$                                         & $0.853$                                       & $0.298$                                         \\
\multicolumn{2}{c}{Lofi}                       & $0.019$                                       & $0.066$                                         & $0.025$                                                      & $0.131$                                                        & $1.000$                                       & $0.133$                                         & $0.767$                                       & $0.078$                                         & $0.994$                                       & $0.311$                                         & $0.837$                                       & $0.432$                                         \\
\multicolumn{2}{c}{Ludwig}                     & $0.012$                                       & $0.072$                                         & $0.019$                                                      & $0.093$                                                        & $0.999$                                       & $0.149$                                         & $0.792$                                       & $0.059$                                         & $0.974$                                       & $0.363$                                         & $0.847$                                       & $0.441$                                         \\
\multicolumn{2}{c}{Mayfair}                    & $0.043$                                       & $0.069$                                         & $0.021$                                                      & $0.145$                                                        & $1.000$                                       & $0.219$                                         & $0.755$                                       & $0.052$                                         & $0.982$                                       & $0.220$                                         & $0.775$                                       & $0.276$                                         \\
\multicolumn{2}{c}{Nashville}                  & $0.006$                                       & $0.040$                                         & $0.015$                                                      & $0.130$                                                        & $0.998$                                       & $0.195$                                         & $0.791$                                       & $0.028$                                         & $0.971$                                       & $0.189$                                         & $0.827$                                       & $0.204$                                         \\
\multicolumn{2}{c}{Ashby}                      & $0.006$                                       & $0.087$                                         & $0.016$                                                      & $0.063$                                                        & $0.996$                                       & $0.200$                                         & $0.790$                                       & $0.055$                                         & $0.956$                                       & $0.339$                                         & $0.839$                                       & $0.406$                                         \\
\multicolumn{2}{c}{Brooklyn}                   & $0.017$                                       & $0.102$                                         & $0.018$                                                      & $0.115$                                                        & $1.000$                                       & $0.186$                                         & $0.794$                                       & $0.043$                                         & $0.763$                                       & $0.327$                                         & $0.846$                                       & $0.253$                                         \\
\multicolumn{2}{c}{Charmes}                    & $0.024$                                       & $0.073$                                         & $0.011$                                                      & $0.102$                                                        & $1.000$                                       & $0.134$                                         & $0.786$                                       & $0.049$                                         & $0.968$                                       & $0.295$                                         & $0.840$                                       & $0.287$                                         \\
\multicolumn{2}{c}{Clarendon}                  & $0.005$                                       & $0.043$                                         & $0.015$                                                      & $0.208$                                                        & $1.000$                                       & $0.120$                                         & $0.785$                                       & $0.068$                                         & $0.989$                                       & $0.373$                                         & $0.822$                                       & $0.270$                                         \\
\multicolumn{2}{c}{Crema}                      & $0.019$                                       & $0.059$                                         & $0.015$                                                      & $0.147$                                                        & $0.995$                                       & $0.137$                                         & $0.793$                                       & $0.041$                                         & $0.914$                                       & $0.260$                                         & $0.833$                                       & $0.142$                                         \\
\multicolumn{2}{c}{Dogpatch}                   & $0.004$                                       & $0.106$                                         & $0.014$                                                      & $0.200$                                                        & $1.000$                                       & $0.154$                                         & $0.776$                                       & $0.087$                                         & $0.988$                                       & $0.331$                                         & $0.000$                                       & $0.334$                                         \\
\multicolumn{2}{c}{Earlybird}                  & $0.010$                                       & $0.039$                                         & $0.012$                                                      & $0.181$                                                        & $1.000$                                       & $0.093$                                         & $0.686$                                       & $0.065$                                         & $0.995$                                       & $0.224$                                         & $0.623$                                       & $0.200$                                         \\
\multicolumn{2}{c}{Gingham}                    & $0.013$                                       & $0.014$                                         & $0.017$                                                      & $0.131$                                                        & $1.000$                                       & $0.126$                                         & $0.782$                                       & $0.046$                                         & $0.890$                                       & $0.128$                                         & $0.746$                                       & $0.266$                                         \\
\multicolumn{2}{c}{Hefe}                       & $0.011$                                       & $0.070$                                         & $0.019$                                                      & $0.139$                                                        & $1.000$                                       & $0.225$                                         & $0.755$                                       & $0.047$                                         & $0.987$                                       & $0.613$                                         & $0.822$                                       & $0.253$                                         \\
\multicolumn{2}{c}{Helena}                     & $0.007$                                       & $0.044$                                         & $0.014$                                                      & $0.152$                                                        & $1.000$                                       & $0.156$                                         & $0.803$                                       & $0.049$                                         & $0.960$                                       & $0.498$                                         & $0.834$                                       & $0.239$                                         \\
\multicolumn{2}{c}{Hudson}                     & $0.006$                                       & $0.063$                                         & $0.017$                                                      & $0.127$                                                        & $1.000$                                       & $0.147$                                         & $0.797$                                       & $0.019$                                         & $0.993$                                       & $0.231$                                         & $0.832$                                       & $0.078$                                         \\
\multicolumn{2}{c}{Inkwell}                    & $0.061$                                       & $0.084$                                         & $0.016$                                                      & $0.053$                                                        & $0.994$                                       & $0.236$                                         & $0.780$                                       & $0.052$                                         & $0.934$                                       & $0.456$                                         & $0.655$                                       & $0.133$                                         \\
\multicolumn{2}{c}{Juno}                       & $0.010$                                       & $0.043$                                         & $0.015$                                                      & $0.113$                                                        & $1.000$                                       & $0.142$                                         & $0.785$                                       & $0.090$                                         & $0.978$                                       & $0.476$                                         & $0.843$                                       & $0.544$                                         \\
\multicolumn{2}{c}{Kelvin}                     & $0.005$                                       & $0.025$                                         & $0.016$                                                      & $0.070$                                                        & $1.000$                                       & $0.118$                                         & $0.781$                                       & $0.058$                                         & $0.997$                                       & $0.187$                                         & $0.803$                                       & $0.509$                                         \\
\multicolumn{2}{c}{Lark}                       & $0.065$                                       & $0.062$                                         & $0.018$                                                      & $0.065$                                                        & $1.000$                                       & $0.156$                                         & $0.795$                                       & $0.050$                                         & $0.960$                                       & $0.408$                                         & $0.845$                                       & $0.130$                                         \\
\multicolumn{2}{c}{Maven}                      & $0.008$                                       & $0.011$                                         & $0.014$                                                      & $0.095$                                                        & $0.996$                                       & $0.077$                                         & $0.809$                                       & $0.032$                                         & $0.874$                                       & $0.096$                                         & $0.835$                                       & $0.115$                                         \\
\multicolumn{2}{c}{Moon}                       & $0.015$                                       & $0.052$                                         & $0.015$                                                      & $0.062$                                                        & $1.000$                                       & $0.107$                                         & $0.784$                                       & $0.036$                                         & $0.965$                                       & $0.244$                                         & $0.726$                                       & $0.163$                                         \\
\multicolumn{2}{c}{Perpetua}                   & $0.011$                                       & $0.020$                                         & $0.015$                                                      & $0.090$                                                        & $1.000$                                       & $0.124$                                         & $0.807$                                       & $0.035$                                         & $0.997$                                       & $0.167$                                         & $0.835$                                       & $0.153$                                         \\
\multicolumn{2}{c}{Poprocket}                  & $0.008$                                       & $0.009$                                         & $0.015$                                                      & $0.081$                                                        & $1.000$                                       & $0.053$                                         & $0.755$                                       & $0.045$                                         & $0.774$                                       & $0.272$                                         & $0.605$                                       & $0.229$                                         \\
\multicolumn{2}{c}{Reyes}                      & $0.007$                                       & $0.032$                                         & $0.016$                                                      & $0.105$                                                        & $1.000$                                       & $0.081$                                         & $0.797$                                       & $0.015$                                         & $0.991$                                       & $0.201$                                         & $0.825$                                       & $0.093$                                         \\
\multicolumn{2}{c}{Rise}                       & $0.002$                                       & $0.030$                                         & $0.006$                                                      & $0.089$                                                        & $1.000$                                       & $0.085$                                         & $0.798$                                       & $0.031$                                         & $0.992$                                       & $0.357$                                         & $0.825$                                       & $0.231$                                         \\
\multicolumn{2}{c}{Sierra}                     & $0.011$                                       & $0.101$                                         & $0.015$                                                      & $0.160$                                                        & $1.000$                                       & $0.202$                                         & $0.785$                                       & $0.087$                                         & $0.964$                                       & $0.326$                                         & $0.775$                                       & $0.210$                                         \\
\multicolumn{2}{c}{Skyline}                    & $0.019$                                       & $0.139$                                         & $0.013$                                                      & $0.230$                                                        & $0.999$                                       & $0.243$                                         & $0.784$                                       & $0.088$                                         & $0.974$                                       & $0.396$                                         & $0.825$                                       & $0.390$                                         \\
\multicolumn{2}{c}{Slumber}                    & $0.001$                                       & $0.027$                                         & $0.005$                                                      & $0.097$                                                        & $1.000$                                       & $0.104$                                         & $0.800$                                       & $0.020$                                         & $0.990$                                       & $0.128$                                         & $0.841$                                       & $0.071$                                         \\
\multicolumn{2}{c}{Stinson}                    & $0.053$                                       & $0.019$                                         & $0.018$                                                      & $0.063$                                                        & $1.000$                                       & $0.068$                                         & $0.793$                                       & $0.015$                                         & $0.990$                                       & $0.125$                                         & $0.790$                                       & $0.082$                                         \\
\multicolumn{2}{c}{Sutro}                      & $0.006$                                       & $0.024$                                         & $0.015$                                                      & $0.117$                                                        & $1.000$                                       & $0.128$                                         & $0.785$                                       & $0.035$                                         & $0.969$                                       & $0.171$                                         & $0.832$                                       & $0.138$                                         \\
\multicolumn{2}{c}{Toaster}                    & $0.027$                                       & $0.070$                                         & $0.019$                                                      & $0.136$                                                        & $1.000$                                       & $0.188$                                         & $0.782$                                       & $0.085$                                         & $0.901$                                       & $0.288$                                         & $0.725$                                       & $0.190$                                         \\
\multicolumn{2}{c}{Valencia}                   & $0.002$                                       & $0.043$                                         & $0.010$                                                      & $0.124$                                                        & $1.000$                                       & $0.147$                                         & $0.808$                                       & $0.022$                                         & $0.994$                                       & $0.175$                                         & $0.837$                                       & $0.115$                                         \\
\multicolumn{2}{c}{Walden}                     & $0.066$                                       & $0.025$                                         & $0.021$                                                      & $0.097$                                                        & $1.000$                                       & $0.081$                                         & $0.787$                                       & $0.034$                                         & $0.924$                                       & $0.244$                                         & $0.794$                                       & $0.329$                                         \\
\multicolumn{2}{c}{Willow}                     & $0.004$                                       & $0.029$                                         & $0.013$                                                      & $0.047$                                                        & $1.000$                                       & $0.070$                                         & $0.784$                                       & $0.037$                                         & $0.915$                                       & $0.073$                                         & $0.712$                                       & $0.066$                                         \\
\multicolumn{2}{c}{XPro2}                      & $0.050$                                       & $0.053$                                         & $0.023$                                                      & $0.179$                                                        & $0.999$                                       & $0.156$                                         & $0.762$                                       & $0.029$                                         & $0.947$                                       & $0.185$                                         & $0.784$                                       & $0.088$                                         \\* \bottomrule
\end{longtable}

% \begin{figure}[!ht]
%     \centering
%     \includegraphics[width=\textwidth]{figs/ensemble-obfs.png}
%     \caption{Outputs of the Ensemble configurations.}
%     \label{fig:ensemble-obfs}
% \end{figure}

\end{document}